\pdfoutput=1
\PassOptionsToPackage{most,breakable}{tcolorbox}
\PassOptionsToPackage{dvipsnames,table}{xcolor}
\PassOptionsToPackage{numbers,compress}{natbib}
\documentclass[11pt]{article}

\usepackage[final]{ropedia}

\usepackage[T1]{fontenc}
\usepackage[utf8]{inputenc}
\usepackage{amsmath,amssymb}
\usepackage{amsfonts}
\usepackage{mathpazo}
\usepackage[xcharter,bigdelims,vvarbb]{newtxmath}
\usepackage[scaled]{helvet}
\usepackage[scaled=1.1]{zlmtt}
\usepackage{latexsym}
\usepackage{pifont}
\usepackage{placeins}
\usepackage{booktabs}
\usepackage{multirow}
\usepackage{tabularx}
\usepackage{subcaption}
\usepackage{arydshln}
\usepackage{adjustbox}
\usepackage{makecell}
\usepackage{enumitem}
\usepackage{rotating}
\usepackage{titletoc}
\usepackage{comment}
\usepackage{wrapfig}
\usepackage{needspace}
\usepackage{nicefrac}
\usepackage{microtype}
\usepackage{siunitx}
\usepackage{dblfloatfix}
\usepackage{graphicx}
\usepackage{cuted}
\usepackage{capt-of}
\usepackage{xspace}
\usepackage{flafter}
\setlist[itemize]{leftmargin=2em}


\usepackage{tikz}
\usetikzlibrary{fadings}
\usepackage{array}
\usepackage{tcolorbox}
\usepackage{threeparttable}

\newcommand{\projecttitle}{Holi-World}
\newcommand{\projectsubtitle}{Building Holistic World Intelligence\\from Human-Ego Experience}

\newcommand{\cmark}{\textcolor{green!60!black}{\ding{51}}}
\newcommand{\xmark}{\textcolor{red}{\ding{55}}}
\definecolor{subcol}{HTML}{F7FFF2}
\colorlet{rankfirst}{ropediagreen}
\colorlet{ranksecond}{ropedialightgreen}
\colorlet{rankthird}{ropediaabstractgreen}
\colorlet{bestgreen}{rankfirst}
\colorlet{secondgreen}{ranksecond}
\colorlet{thirdgreen}{rankthird}
\colorlet{bestred}{rankfirst}
\colorlet{secondorange}{ranksecond}
\colorlet{thirdyellow}{rankthird}
\definecolor{trainmark}{RGB}{0,180,0}
\definecolor{oomred}{RGB}{252,228,228}
\definecolor{catgray}{HTML}{E7F8DC}
\definecolor{altcolgreen}{HTML}{F0FFE8}
\colorlet{altcolblue}{altcolgreen}
\definecolor{pretrainmark}{RGB}{197,90,17}
\definecolor{gaingreen}{RGB}{30,140,60}
\definecolor{lossred}{RGB}{200,30,30}
\definecolor{navyblue}{HTML}{4F8F35}

\colorlet{findingback}{ropedialightgreen}
\colorlet{findingbacksoft}{ropediaabstractgreen}
\colorlet{findingframe}{ropediadeepgreen}
\colorlet{findingaccent}{ropediaaccent}

\newcommand{\best}[1]{\cellcolor{rankfirst}\textbf{#1}}
\newcommand{\second}[1]{\cellcolor{ranksecond}#1}
\colorlet{fullconfig}{ropedialightgreen}
\newcommand{\rankswatch}[1]{{\setlength{\fboxsep}{0pt}\setlength{\fboxrule}{0.4pt}%
  \fcolorbox{black!35}{#1}{\rule{0pt}{0.7em}\rule{1.1em}{0pt}}}}

\firstpagelogos{logo/ropedia_Green.png}{logo/ntu_logo.jpg}{}
\renewcommand{\firstpagelogoscale}{1.5}
\abstractlogo{logo/Horizontal_Black.png}

\title{\projecttitle}
\subtitle{\projectsubtitle}

\newcommand{\HWAuthor}[2]{\mbox{\textbf{#1}\textsuperscript{\textcolor{ropediaaccent}{#2}}}}
\newcommand{\HWAffil}[2]{\mbox{\textsuperscript{\textcolor{ropediaaccent}{#1}}#2}}
\newcommand{\HWNote}[2]{\mbox{\textsuperscript{\textcolor{ropediaaccent}{#1}}\textcolor{black!55}{#2}}}

\author{
  \HWAuthor{Hao Li}{1,2*},
  \HWAuthor{Xiyuan Zheng}{1,2*},
  \HWAuthor{Yuanyuan Gao}{1,3*},
  \HWAuthor{Yalun Dai}{1,2},
  \HWAuthor{Xinhao Ji}{1,4},
  \HWAuthor{Xuan-Li Chen}{1},
  \HWAuthor{Jing Zhao}{1},
  \HWAuthor{Fangzhou Hong}{1},
  \HWAuthor{Zhaoxi Chen}{1},
  \HWAuthor{Ziwei Liu}{1,2$\dagger$}\\[1.0ex]
  {\normalfont\footnotesize
  \HWAffil{1}{Ropedia}\qquad
  \HWAffil{2}{NTU}\qquad
  \HWAffil{3}{HKUST}\qquad
  \HWAffil{4}{PKU}\qquad
  \HWNote{*}{Equal contribution}\qquad
  \HWNote{$\dagger$}{Corresponding author}}
}

\begin{document}
\maketitle

\begin{paperresources}
\paperresourceicon{\resourcegithubicon}{Code}{https://github.com/Ropedia/holi-world}{github.com/Ropedia/holi-world}
\quad
\paperresourceicon{\resourcehficon}{Model}{https://huggingface.co/ropedia-ai/holiworld-vl}{huggingface.co/ropedia-ai/holiworld-vl}\\
\paperresourceicon{\resourcehficon}{Dataset}{https://huggingface.co/datasets/ropedia-ai/holiworld-data}{huggingface.co/datasets/ropedia-ai/holiworld-data}
\end{paperresources}

\vspace{6pt}
\begin{abstract}
Human egocentric experience offers a rich foundation for learning how the world
is organized and how it changes through interaction. Realizing this potential
at scale requires turning growing volumes of recordings into learning resources
that connect scene structure, object semantics, and interaction histories
across viewpoints and time, while reducing dependence on costly manual
annotation. We introduce \textbf{\textsc{HoliWorld}}, an integrated ecosystem
connecting experience capture, world representation, automated data
construction, model learning, and independent evaluation. To support learning
at scale, \textbf{\textsc{HoliWorld}-\textit{Foundry}} combines complementary static
and dynamic perception to automatically construct world records that connect
persistent scene structure and evolving object states with interaction
histories and supporting observations. \textbf{\textsc{HoliWorld}-\textit{Data}}
organizes these records from approximately 100 self-collected scenes into
grounded learning resources, including approximately 80K QA pairs used to train
\textbf{\textsc{HoliWorld}-\textit{VL-4B}}. Independent human annotations of held-out
recordings form \textbf{\textsc{HoliWorld}-\textit{Bench}}, supporting evaluation of
both annotation quality and learned spatial and temporal understanding.
Experiments on ScanNet, ScanNet++, Aria Digital Twin, and \textsc{HoliWorld}-\textit{Bench} demonstrate strong performance in static perception and dynamic tracking, while training on \textsc{HoliWorld}-\textit{Data} substantially improves \textsc{HoliWorld}-\textit{VL-4B}'s spatial and temporal reasoning on \textsc{HoliWorld}-\textit{Bench} and EgoDynamic4D.
\textsc{HoliWorld} provides a practical foundation for expanding learning
resources from the continuous flow of human experience and advancing machine
understanding of the dynamic world.

\end{abstract}

\begin{figure}[t]
\centering
\includegraphics[width=\textwidth]{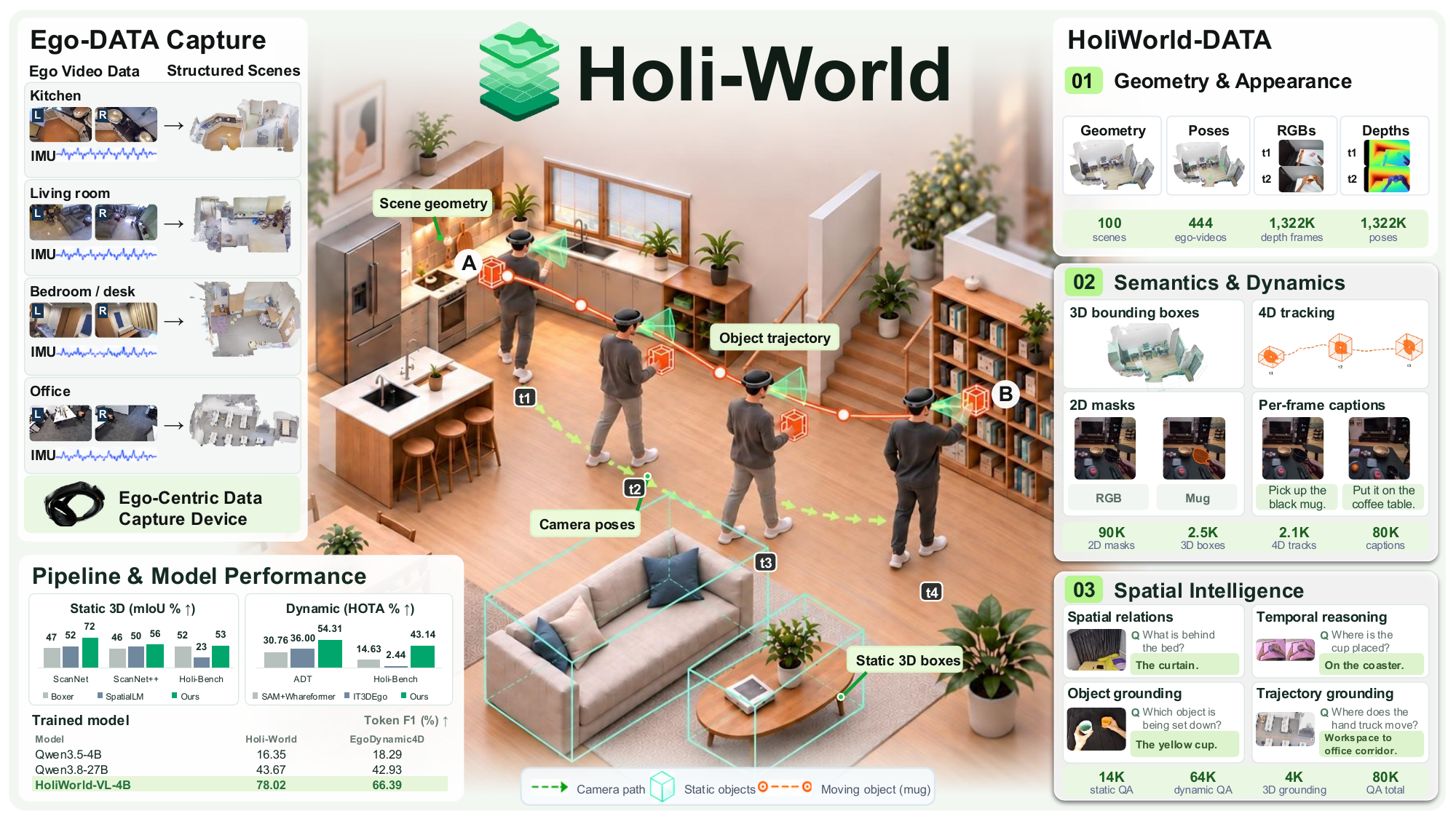}
\caption{\textbf{Overview of \textsc{HoliWorld}.} \textsc{HoliWorld} connects in-the-wild egocentric capture, automated static and dynamic annotation, and vision-language learning. Scene geometry, object semantics, and interaction histories form aligned multimodal supervision for grounding and spatial and temporal reasoning. Representative annotations and results illustrate data coverage, annotation quality, and \textsc{HoliWorld}-\textit{VL} performance.}
\label{fig:teaser}
\end{figure}

\section{Introduction}
\label{sec:introduction}

Learning how the physical world is organized and how it changes through
interaction is a fundamental goal of artificial intelligence and an essential
step toward artificial general intelligence
(AGI)~\citep{tenenbaum2011grow,lake2017machines,lecun2022path}.
Although vision--language learning and predictive video modeling enable
machines to recognize objects, describe scenes, interpret activities, and
anticipate visual
changes~\citep{radford2021clip,li2023blip2,liu2023llava,bai2025qwen25vl,assran2025vjepa2},
building a coherent understanding of the world across viewpoints and time
remains an open challenge~\citep{yang2025vsi,wang2026mindcube,lin2025mmsivideo}.
Everyday human experience naturally brings these dimensions together: moving
through an environment reveals its enduring spatial structure, while
interacting with objects reveals how they are used and how their states and
relationships change. 
Therefore, we place human experience at the center of
learning world intelligence, asking how it should be captured, organized, and
learned from to support an understanding of environments, objects, and
interactions as parts of a shared, evolving world.

Egocentric data offers a direct route to learning about the world through the
breadth of human life. It records environments as people experience them,
bringing together the places they inhabit, the objects they use, and the
activities through which their surroundings
change~\citep{ragusa2021meccano,damen2022rescaling,tschernezki2023epicfields}.
Large-scale efforts~\citep{grauman2022ego4d,grauman2024egoexo4d,ma2024nymeria,yang2025egolife}
have expanded this record across diverse people, places, and activities.
However, turning these recordings into detailed learning supervision often
requires substantial human input, tying further expansion to additional
annotation and curation costs~\citep{darkhalil2022visor}. The challenge is
therefore to scale high-quality supervision with incoming experience while
retaining the spatial and temporal information needed to understand
environments, objects, and interactions as parts of an evolving world.

Recent advances provide complementary capabilities for automating this
process: reconstruction and semantic annotation of predominantly static scenes
for spatial supervision~\citep{holispatial2026}, and egocentric tracking of
object identities and 3D locations over
time~\citep{zhao2024it3dego,plizzari2025osnom}. Turning these capabilities into
coherent learning resources requires connecting scene geometry and object
semantics with interaction histories, then deriving supervision that remains
traceable to the underlying observations. To this end, we introduce
\textbf{\textsc{HoliWorld}}, an integrated ecosystem for automatically
constructing high-quality learning resources at scale from egocentric
recordings that we collect in the wild using our capture device. At its core,
\textbf{\textsc{HoliWorld}-\textit{Foundry}} first reconstructs scene geometry and camera
poses, then uses VLM-derived motion and interaction cues to separate static
objects from dynamic tracking targets. For static objects, it lifts image
detections into 3D and consolidates them with scene point-cloud detections to
recover object geometry and semantics. For dynamic objects, it associates
identities across interaction segments, tracks object regions in image space,
and recovers 3D states and trajectories in shared world coordinates. Built on
these multimodal records, \textbf{\textsc{HoliWorld}-\textit{Data}} organizes
approximately 100 scenes and 444 egocentric videos across four
levels: raw observations, reconstructed geometry, object and interaction
semantics, and spatial intelligence supervision. By further parsing geometric
relations, object trajectories, and interaction states, we construct approximately
80K evidence-driven QA pairs covering spatial relations, temporal
changes, and interactions, with each answer linked to its supporting
observations and annotations. We train \textbf{\textsc{HoliWorld}-\textit{VL}} on this
supervision for visual grounding and spatial reasoning, providing a downstream
test of the generated data's learning value. Finally, we establish
\textbf{\textsc{HoliWorld}-\textit{Bench}} through independent human annotation of
held-out recordings, providing reference annotations for evaluating both
Foundry's world-record accuracy and \textsc{HoliWorld}-\textit{VL}'s spatial understanding.

Experiments evaluate \textsc{HoliWorld} through both annotation quality and downstream
spatial reasoning. For static perception,
\textsc{HoliWorld}-\textit{Foundry} achieves an AP$_{25}$ of 0.68 on
ScanNet~\citep{scannet}, compared with 0.40 for the strongest baseline
in our evaluation. For dynamic tracking, it achieves a HOTA$_{0.1}$ of
54.31\% on 20 Aria Digital Twin sequences~\citep{aria_digital_twin},
compared with 36.00\% for IT3DEgo~\citep{zhao2024it3dego} with
ground-truth-guided initialization. The resulting supervision also supports
substantial improvements in downstream reasoning: compared with the original
Qwen3.5-4B, \textbf{\textsc{HoliWorld}-\textit{VL-4B}} improves Token F1 from
16.35\% to 78.02\% on \textsc{HoliWorld}-\textit{Bench} and from
18.29\% to 66.39\% on
EgoDynamic4D~\citep{huang2026egodynamic4d}.

Our contributions are threefold:
\begin{itemize}
    \item We introduce \textbf{\textsc{HoliWorld}}, an integrated ecosystem that
    provides a practical blueprint for learning world intelligence from human
    egocentric experience. It connects experience capture, world representation,
    automated data construction, model learning, and independent evaluation
    around a shared representation of the evolving world.
    \item We develop \textbf{\textsc{HoliWorld}-\textit{Foundry}} to automatically construct
    world records that preserve persistent scene structure and evolving
    object states. Complementary static and dynamic perception links geometry,
    semantics, and interaction histories through persistent identities and
    supporting observations, enabling scalable construction of grounded
    supervision.
    \item We construct \textbf{\textsc{HoliWorld}-\textit{Data}} to support grounding and
    spatial and temporal reasoning, and use these resources to train
    \textbf{\textsc{HoliWorld}-\textit{VL}}. We further establish
    \textbf{\textsc{HoliWorld}-\textit{Bench}} through independent human annotation,
    providing reference data for evaluating both the quality of automatically
    constructed world records and the capabilities of learned models.
\end{itemize}

\section{Related Work}
\label{sec:related_work}
\label{app:related_work}

\subsection{Egocentric Datasets and Annotations}

Large-scale egocentric datasets cover diverse human activities and
environments~\citep{damen2022rescaling,grauman2022ego4d,grauman2024egoexo4d}.
EPIC Fields and Aria Digital Twin add camera and scene
geometry~\citep{tschernezki2023epicfields,aria_digital_twin}, while
HD-EPIC~\citep{perrett2025hdepic} connects actions, object movements, and
gaze to 3D digital twins through detailed manual annotation. However,
extending such detailed supervision to new recordings entails further
annotation effort~\citep{darkhalil2022visor,perrett2025hdepic}.
\textsc{HoliWorld} addresses this scaling bottleneck by automatically
recovering scene structure, object states, and interaction annotations from
newly captured recordings and converting them into learning supervision,
reducing dependence on manual labeling as new experience is collected.

\subsection{3D Scene Reconstruction and Object Tracking}

Holi-Spatial~\citep{holispatial2026} automates static scene reconstruction,
semantic annotation, and spatial QA generation at scale, while
Boxer~\citep{boxer2026} consolidates multi-view detections into static 3D
boxes. However, static scene consolidation does not retain the successive
object states needed to describe motion and interactions. Egocentric tracking
provides complementary capabilities: IT3DEgo and
EgoSeg3D~\citep{zhao2024it3dego,bhalgat2024egoseg3d} maintain object identities
and locations over time, and LMK, introduced for the Out of Sight, Not Out of Mind
(OSNOM) task~\citep{plizzari2025osnom}, maintains spatial memory beyond the current view.
\textsc{HoliWorld}-\textit{Foundry} connects these static and dynamic
perception objectives within an automated data construction pipeline. Shared
world coordinates and persistent identities link scene geometry, evolving
object states, and interaction evidence, enabling spatial and temporal
supervision to refer to the same objects and supporting observations.

\subsection{Spatial and Temporal Reasoning in Vision--Language Models}

Dedicated spatial supervision~\citep{chen2024spatialvlm,ma2025spatialllm}
and geometry-aware visual
representations~\citep{zhu2024llava3d,zheng2025video3dllm,wu2025spatialmllm}
improve spatial reasoning, while video
benchmarks~\citep{yang2025vsi,lin2025mmsivideo} test these capabilities across
observations. ST-Think and EgoReasoner~\citep{wu2026stthink,zhu2026egoreasoner}
further study reasoning over egocentric spatial and temporal information.
EgoDynamic4D~\citep{huang2026egodynamic4d} constructs dynamic QA from existing
annotated 4D resources and introduces a model integrating static and dynamic
scene information. We follow its QA patterns, while addressing the upstream
challenge of automatically constructing the underlying annotations from new
recordings. \textsc{HoliWorld} connects experience capture, automated
annotation, supervision generation, and model training, allowing new
recordings to expand the learning resources. Independent human-annotated
evaluation then assesses both the accuracy of the recovered records and the
capabilities learned from them.

\section{Method}
\label{sec:method}

\subsection{HoliWorld Overview}
\label{sec:holiworld_overview}

\begin{figure}[t]
    \centering
    \includegraphics[width=\textwidth]{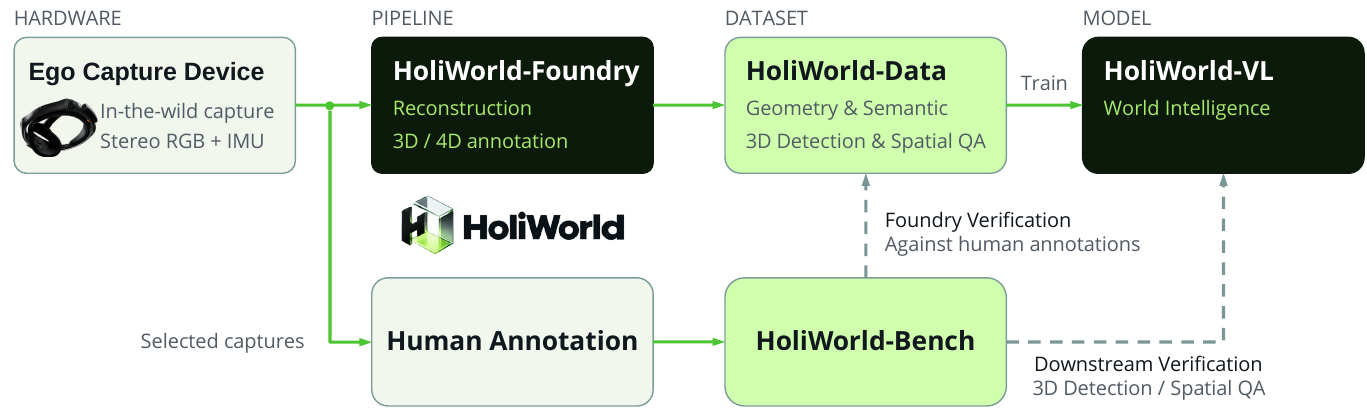}
    \caption{\textbf{The \textsc{HoliWorld} ecosystem.} It connects egocentric
    capture, Foundry, Data, and VL. Human-annotated Bench evaluates annotation
    quality and spatial understanding.}
    \label{fig:overview}
\end{figure}

HoliWorld converts egocentric recordings into spatially and temporally aligned
supervision (Fig.~\ref{fig:overview}). The Eagle Capture Device records
synchronized stereo RGB and IMU measurements; HoliWorld-Foundry reconstructs scenes and annotates
static objects and dynamic tracks. HoliWorld-Data organizes these outputs into
grounding, captioning, and spatial QA supervision for HoliWorld-VL.
Independent human annotations of held-out recordings form HoliWorld-Bench
for evaluating Foundry's annotation quality and HoliWorld-VL's spatial
understanding.

\subsection{HoliWorld Foundry}
\label{sec:foundry}

\begin{figure}[t]
    \centering
    \includegraphics[width=\textwidth]{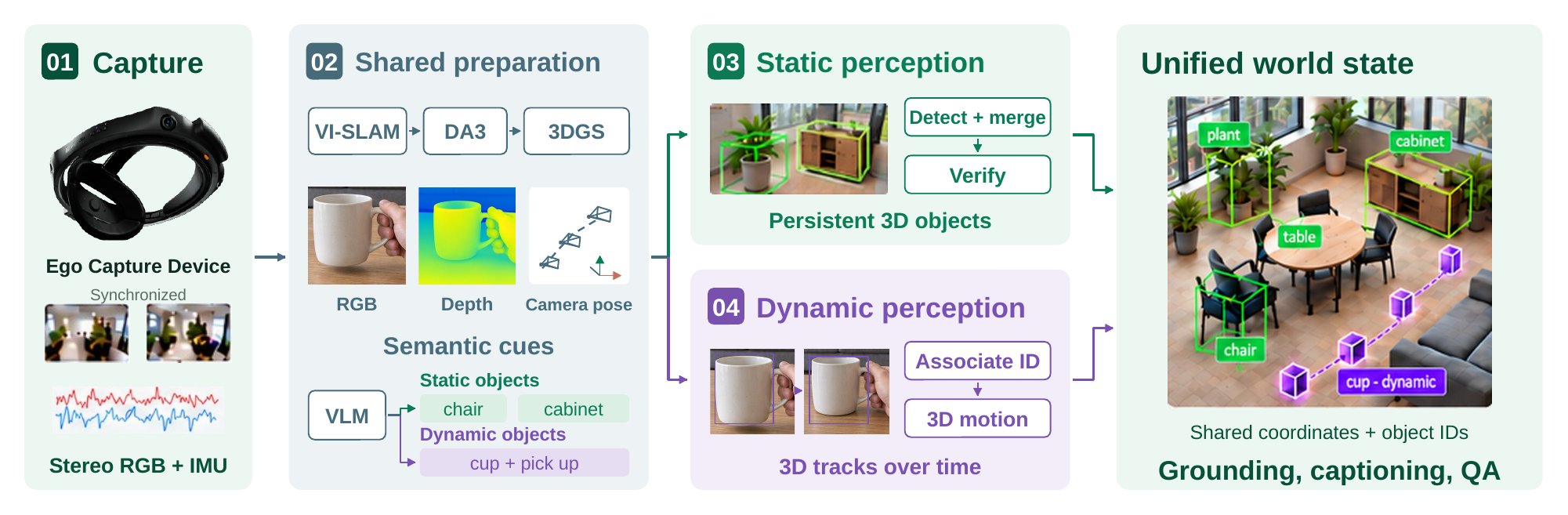}
    \caption{\textbf{HoliWorld-Foundry overview.} Shared preparation reconstructs
    geometry and routes objects to static or dynamic perception. Static
    perception (03) detects, merges, and verifies persistent 3D objects; dynamic
    perception (04) associates identities and recovers 3D tracks. Both outputs
    are fused into a unified world state.}
    \label{fig:framework}
\end{figure}

\subsubsection{Geometric Preparation and Shared Semantic Cues}
\label{sec:geometry_semantics}

As illustrated in Fig.~\ref{fig:framework}, we first recover camera poses and
sparse points from stereo RGB and IMU using
visual-inertial SLAM, then estimate depth and dense points from RGB keyframes
and these poses with Depth Anything 3 (DA3)~\citep{depthanything3}.
Using RGB observations and this geometry, we optimize a 3D Gaussian scene
representation~\cite{kerbl20233d} $\mathcal{G}$, render refined depth at the observed poses, and
complete dynamic regions with per-frame DA3 depth estimation. We retain
$\mathcal{G}$ and the reconstructed scene point cloud $\mathcal{P}$, and
organize the resulting observations as
\begin{equation}
    \mathcal{O}=\{(I_t,D_t,\mathbf{K}_t,\mathbf{T}_t)\}_{t=1}^{T},
    \label{eq:observations}
\end{equation}
where $I_t$ is an RGB frame, $D_t$ is a depth map,
$\mathbf{K}_t$ is the camera intrinsic matrix, and $\mathbf{T}_t$ maps camera
coordinates to world coordinates. To route static and dynamic objects to their
respective perception pipelines, a vision--language model (VLM) extracts
categories, attributes, motion cues, and interaction descriptions from RGB
keyframes in $\mathcal{O}$;
categories guide static image-based proposals, while motion and interaction
cues identify dynamic tracking targets.

\subsubsection{Static Object Perception}
\label{sec:static_foundry}

\begin{figure}[t]
    \centering
    \includegraphics[width=\textwidth]{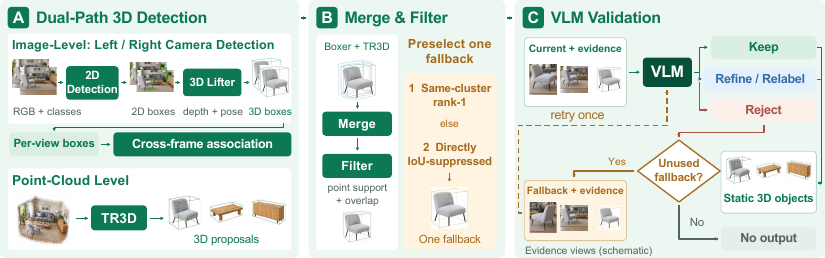}
    \caption{\textbf{Static object perception.} (A) Complementary image and
    point-cloud paths generate 3D candidates. (B) Geometric merging and
    filtering retain a current candidate and one preselected fallback. (C) VLM
    validation uses each candidate's own evidence, with at most one fallback
    retry. Synthetic illustrations depict inputs and intermediate states.}
    \label{fig:foundry_static}
\end{figure}

\begin{figure}[t]
    \centering
    \includegraphics[width=\textwidth]{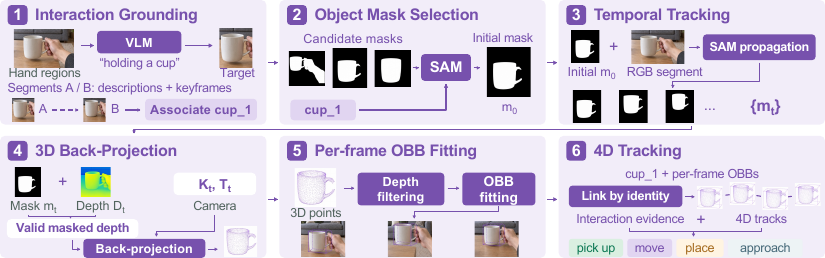}
    \caption{\textbf{Dynamic object perception.} Cross-segment identity
    association guides mask selection and within-segment propagation. Valid
    masked depth yields filtered 3D points and per-frame 3D oriented bounding
    boxes (OBBs). Linking boxes by identity produces 4D tracks; tracks and
    interaction evidence jointly support motion cues. Synthetic illustrations
    depict inputs and intermediate states.}
    \label{fig:foundry_dynamic}
\end{figure}

\paragraph{Dual-path 3D detection.}
To capture both object details and scene-level geometry, we combine image-based
and point-cloud-based 3D detection (Fig.~\ref{fig:foundry_static}A). In the
image-based path, we run Boxer~\citep{boxer2026} separately on the left and
right RGB streams to capture complementary object details. Open-vocabulary 2D
detections are lifted into metric 3D boxes using the corresponding depth and
camera poses, then associated across frames. In parallel, TR3D~\citep{rukhovich2023tr3d} detects objects
in the reconstructed scene point cloud $\mathcal{P}$, capturing large-object
extents beyond individual views.

\paragraph{Box merging and geometric filtering.}
Across these complementary sources, multiple proposals may refer to the same
object, while others have little geometric support. We therefore use low-cost
geometric checks to merge spatially compatible boxes and discard those with
insufficient point support in $\mathcal{P}$ before VLM validation
(Fig.~\ref{fig:foundry_static}B). For validation, we also preserve
one fallback per selected box when available: the rank-1 alternative in its
candidate cluster, or, if unavailable, a directly suppressed candidate linked
by IoU suppression. Each box retains its own supporting evidence.

\paragraph{VLM validation.}
The remaining boxes may still have inaccurate extents or incorrect category
labels. A VLM therefore checks each selected box against source observations
and supplementary views rendered from $\mathcal{G}$
(Fig.~\ref{fig:foundry_static}C). Based on this evidence, it can keep the box,
refine its geometry, relabel its category, or reject it. If the selected box
is rejected, we evaluate its preselected fallback once using the fallback's
own evidence. If no unused fallback is available or the retry fails, no box
is retained. Accepted boxes form $\mathcal{B}^{\mathrm{sta}}$ with links to
their supporting observations.

\subsubsection{Dynamic Object Perception}
\label{sec:dynamic_foundry}

\paragraph{Interaction grounding and identity association.}
Manipulated objects may disappear from view and reappear in later interaction
segments. To recognize the same object across these interruptions, we use
hand regions and VLM interaction descriptions to identify the target in each
segment, then compare descriptions and keyframe evidence across segments.
This \emph{temporal label continuity} establishes consistent object identities
that guide target mask selection (Fig.~\ref{fig:foundry_dynamic}).

\paragraph{Identity-guided mask tracking.}
The associated identities specify which object to follow, while frame-level
masks are needed to localize it over time. Guided by these identities, the
Segment Anything Model (SAM)~\citep{kirillov2023sam,ravi2025sam2} selects and initializes the target mask, then propagates it within each
interaction segment to obtain $\{m_t\}$.
We assume at most one moving object at each time, with $m_t$ denoting its
foreground pixels.

\paragraph{3D recovery and motion cues.}
Image-space motion also reflects camera movement. To recover object states
in world coordinates, we back-project the pixels of each valid tracked mask
using the DA3-completed dynamic depths in $D_t$ and the camera pose:
\begin{equation}
\mathcal{X}_t=
\left\{
\mathbf{T}_t
\left(D_t(\mathbf{u})\mathbf{K}_t^{-1}\bar{\mathbf{u}}\right)
\;\middle|\;
\mathbf{u}\in m_t,\;D_t(\mathbf{u})\ \text{is valid}
\right\},
\label{eq:dynamic_backprojection}
\end{equation}
where $\bar{\mathbf{u}}$ is the homogeneous image coordinate,
$\mathbf{T}_t(\cdot)$ applies the camera-to-world rigid transformation, and
$\mathcal{X}_t$ contains the object-supported world points.
After filtering unsupported depth samples, we fit an OBB to each
point set and link boxes with the same identity across time to form a dynamic
4D track.

\subsubsection{Unified World Representation}

Object records connect instance identities and categories to image evidence,
3D geometry, and interaction cues. We express the Foundry output at time $t$
as
\begin{equation}
\mathcal{S}(t)=
\big(
    \mathcal{O}_t,\mathcal{U}(t),
    \mathcal{G},\mathcal{P},
    \mathcal{B}^{\mathrm{sta}},\mathcal{B}^{\mathrm{dyn}}(t),
    \mathcal{Z}(t)
\big),
\label{eq:world_state}
\end{equation}
where $\mathcal{O}_t=(I_t,D_t,\mathbf{K}_t,\mathbf{T}_t)$ is the input
observation, $\mathcal{U}(t)$ contains the detected 2D boxes and tracked masks,
and $\mathcal{Z}(t)$ records the associated labels, identities, confidence scores, interaction
semantics, and evidence links. Static instances persist across observations;
dynamic instances retain time-indexed states. This representation lets
supervision refer to specific objects, times, and supporting observations.

\subsection{HoliWorld-Data}
\label{sec:data_qa}

\begin{wrapfigure}{r}{0.48\textwidth}
  \vspace{-1.0\baselineskip}
  \centering
  \includegraphics[width=\linewidth]{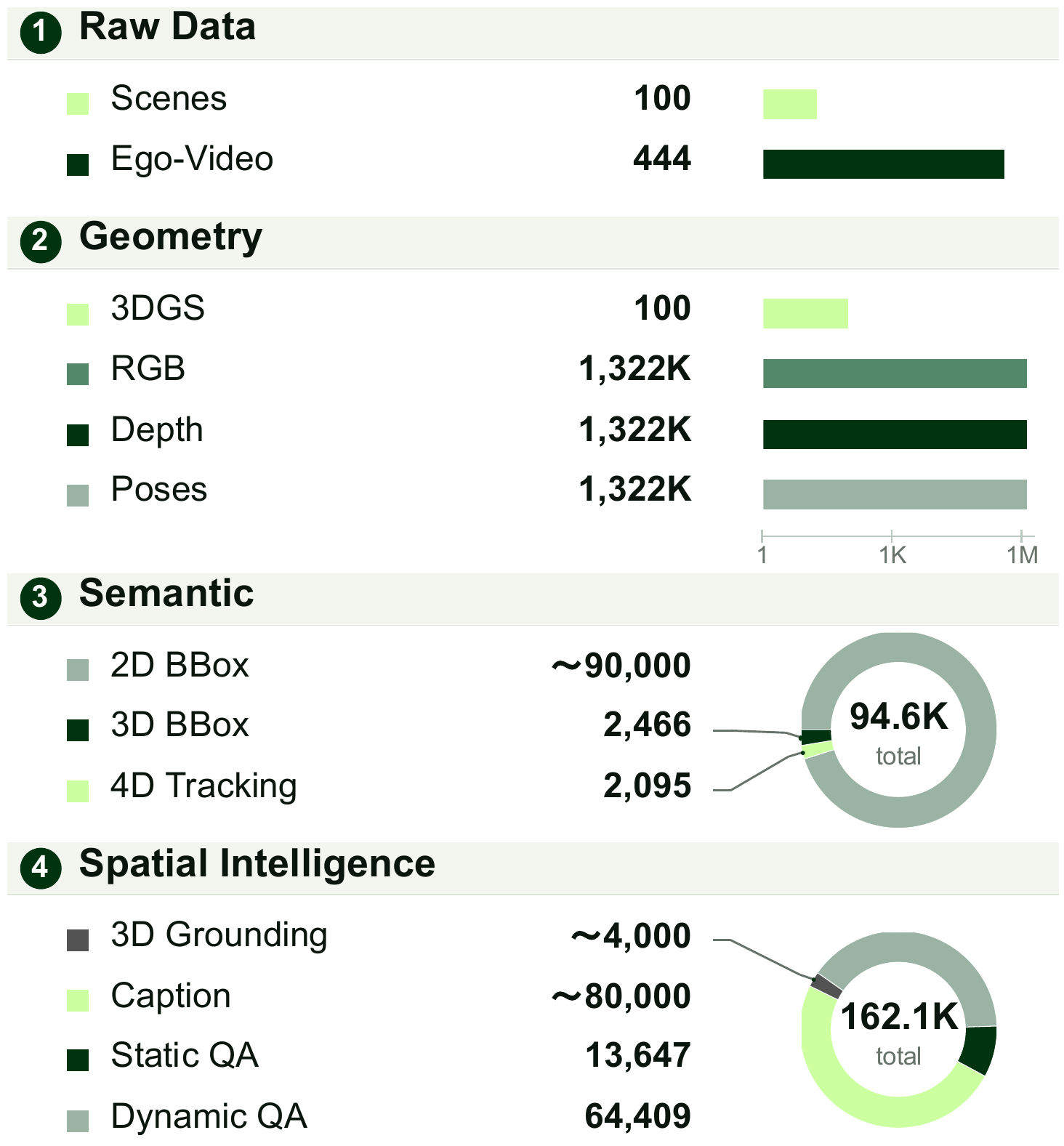}
  \caption{\textbf{HoliWorld-Data composition} and its spatial-temporal
  supervision.}
  \label{fig:holiworld_data}
  \vspace{-0.5\baselineskip}
\end{wrapfigure}

HoliWorld-Data organizes the world records constructed by Foundry into
grounding, captioning, and spatial, temporal, and interaction QA supervision.
The available evidence determines the scope of these tasks: static objects
and their geometry support spatial understanding; dynamic tracks with
consistent object identities support motion and temporal understanding; and
tracks paired with interaction semantics support descriptions of object
manipulation and state changes. Each example retains links to source
observations and annotations, allowing its target output to be traced to
specific objects, times, and supporting evidence.

\textbf{QA pairs construction.}
We construct QA pairs from the spatial, temporal, and interaction annotations
in Foundry's world records, following the QA patterns of
EgoDynamic4D~\citep{huang2026egodynamic4d}.
Grounding tasks associate referring expressions with the corresponding 2D
observations, static 3D boxes, or dynamic tracks. Spatial QA uses object
geometry and camera poses to derive direction, distance, containment,
support, and viewpoint-dependent relations. Temporal QA compares the states
of the same object over time to describe motion, event ordering, and duration.
Interaction QA additionally uses interaction semantics and source observations
to describe manipulated objects and state changes. Scene captions and
inventory questions aggregate object records across frames to describe the
observed environment. Dynamic supervision follows Foundry's current
assumption of at most one moving object at each time.

\textbf{Dataset composition.}
HoliWorld-Data contains roughly 100 scenes and 444 egocentric videos,
with approximately 100 3D Gaussian splatting (3DGS) reconstructions,
1,322K RGB frames with corresponding depth maps and camera poses,
90K 2D boxes, 2,466 static 3D boxes, and 2,095 dynamic tracks
(Fig.~\ref{fig:holiworld_data}). These records support about 4K 3D grounding
annotations and 80K captions, together with 13,647 static QA pairs and 64,409
dynamic QA pairs, totaling about 80K QA pairs.

\FloatBarrier
\section{Experiments}
\label{sec:experiments}

We evaluate the pipeline's ability to generate object-centric 3D annotations
for both static scenes and moving objects.
Sections~\ref{sec:static_3d_bbox} and~\ref{sec:adt_dynamic_tracking} present
the results for static object perception and
dynamic object tracking.
Section~\ref{sec:ablation_study} examines the contributions of key pipeline
components through controlled ablations. Detailed experimental settings are provided in Appendix Sections~\ref{app:static_evaluation} and~\ref{app:dynamic_full_metrics}.

\subsection{Static 3D Object Perception}
\label{sec:static_3d_bbox}

\begin{figure}[htbp]
    \centering
    \includegraphics[width=\textwidth]{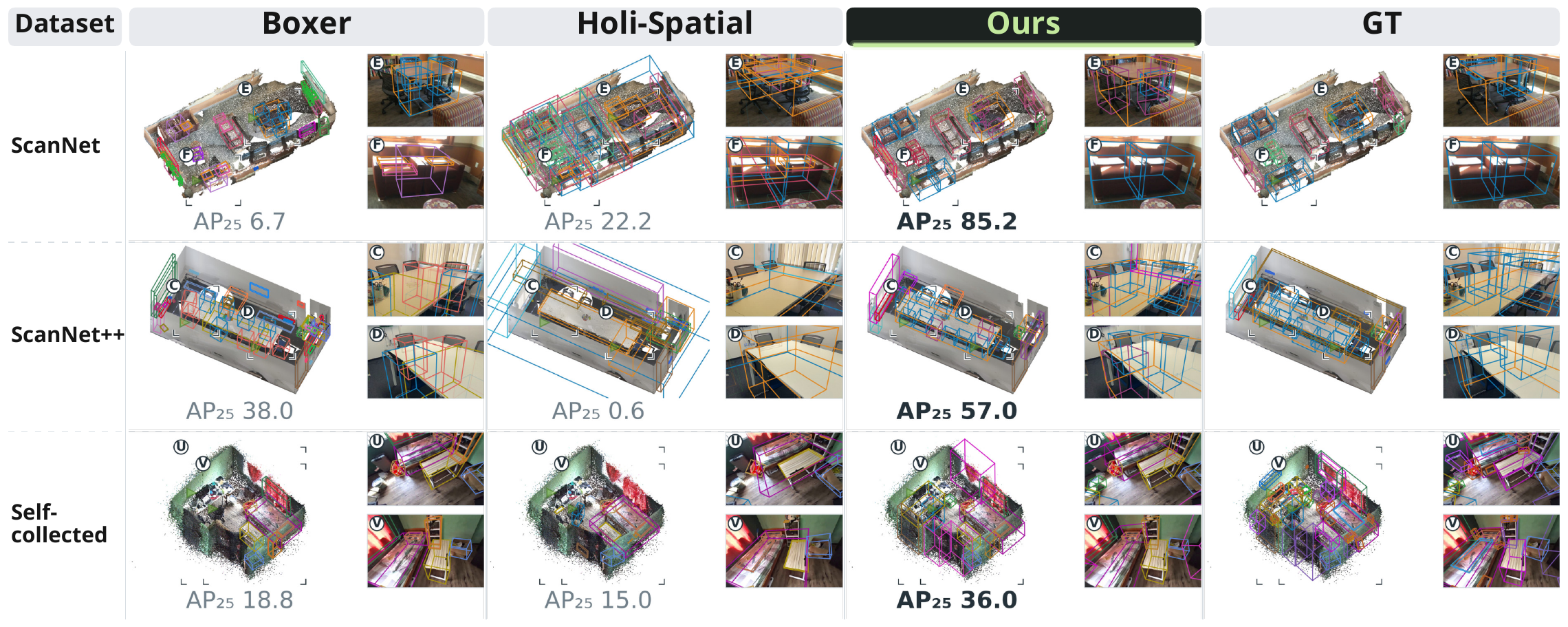}
    \caption{\textbf{Qualitative static 3D object perception results} across
    representative scenes.}
    \label{fig:static_main_results}
\end{figure}

As shown in Table~\ref{tab:static_3d_bbox}, \textsc{HoliWorld-Foundry} achieves the highest AP$_{25}$ and AP$_{50}$ on all three datasets. On ScanNet~\citep{scannet} and ScanNet++~\citep{scannetpp}, AP$_{50}$ reaches 0.55 and 0.13, respectively, compared with 0.12 and 0.05 for the strongest baselines, indicating more accurate detection under stricter geometric matching. The method also leads on the more complex real-world scenes in HoliWorld-Bench. Fig.~\ref{fig:static_main_results} shows more complete object coverage and better-aligned box extents across representative scenes, consistent with these quantitative gains.

\begin{table}[htbp]
\centering
\caption{\textbf{Static 3D perception} on ScanNet, ScanNet++, and HoliWorld-Bench
($[0,1]$ scale). AP uses 3D IoU thresholds of 0.25/0.50; mIoU uses pairs
matched at IoU 0.25. Best and second-best results are highlighted in
deep green~\rankswatch{rankfirst} and light green~\rankswatch{ranksecond}.}
\label{tab:static_3d_bbox}
\small
\setlength{\tabcolsep}{3pt}
\renewcommand{\arraystretch}{1.12}
\begin{tabular*}{\textwidth}{@{\extracolsep{\fill}}lccccccccc@{}}
\toprule
& \multicolumn{3}{c}{\textbf{ScanNet}} & \multicolumn{3}{c}{\textbf{ScanNet++}}
& \multicolumn{3}{c}{\textbf{HoliWorld-Bench}} \\
\cmidrule(lr){2-4}\cmidrule(lr){5-7}\cmidrule(lr){8-10}
\textbf{Method}
& AP$_{25}\uparrow$ & AP$_{50}\uparrow$ & mIoU$\uparrow$
& AP$_{25}\uparrow$ & AP$_{50}\uparrow$ & mIoU$\uparrow$
& AP$_{25}\uparrow$ & AP$_{50}\uparrow$ & mIoU$\uparrow$ \\
\midrule
\multicolumn{10}{@{}l}{\textit{Model-based methods}} \\
SpatialLM   & 0.18 & 0.09 & \second{0.52} & 0.03 & 0.01 & 0.50 & 0.04 & 0.00 & 0.23 \\
LLaVA-3D    & 0.02 & 0.00 & 0.34 & 0.00 & 0.00 & 0.34 & 0.00 & 0.00 & 0.00 \\
SceneScript & 0.00 & 0.00 & 0.00 & 0.00 & 0.00 & 0.00 & 0.00 & 0.00 & 0.00 \\
Map-Det3D   & \second{0.40} & \second{0.12} & 0.48 & 0.18 & \second{0.05} & 0.48 & 0.12 & 0.01 & 0.37 \\
\midrule
\multicolumn{10}{@{}l}{\textit{3DGS-based methods}} \\
M3-Spatial  & 0.00 & 0.00 & 0.37 & 0.00 & 0.00 & \best{0.60} & 0.06 & 0.00 & 0.28 \\
LangSplat   & 0.02 & 0.00 & 0.45 & 0.01 & 0.00 & 0.33 & 0.01 & 0.00 & 0.10 \\
\midrule
\multicolumn{10}{@{}l}{\textit{Spatial pipeline frameworks}} \\
Boxer        & 0.32 & 0.09 & 0.47 & \second{0.23} & \second{0.05} & 0.46 & \second{0.16} & \second{0.07} & \second{0.52} \\
Holi-Spatial & 0.33 & 0.10 & 0.47 & 0.17 & \second{0.05} & 0.45 & 0.07 & 0.01 & 0.37 \\
\textbf{Ours} & \best{0.68} & \best{0.55} & \best{0.72} & \best{0.30} & \best{0.13} & \second{0.56} & \best{0.24} & \best{0.12} & \best{0.53} \\
\bottomrule
\end{tabular*}
\end{table}

\subsection{Dynamic 3D Object Perception and Tracking}
\label{sec:adt_dynamic_tracking}

\begin{figure}[htbp]
    \centering
    \includegraphics[width=\textwidth]{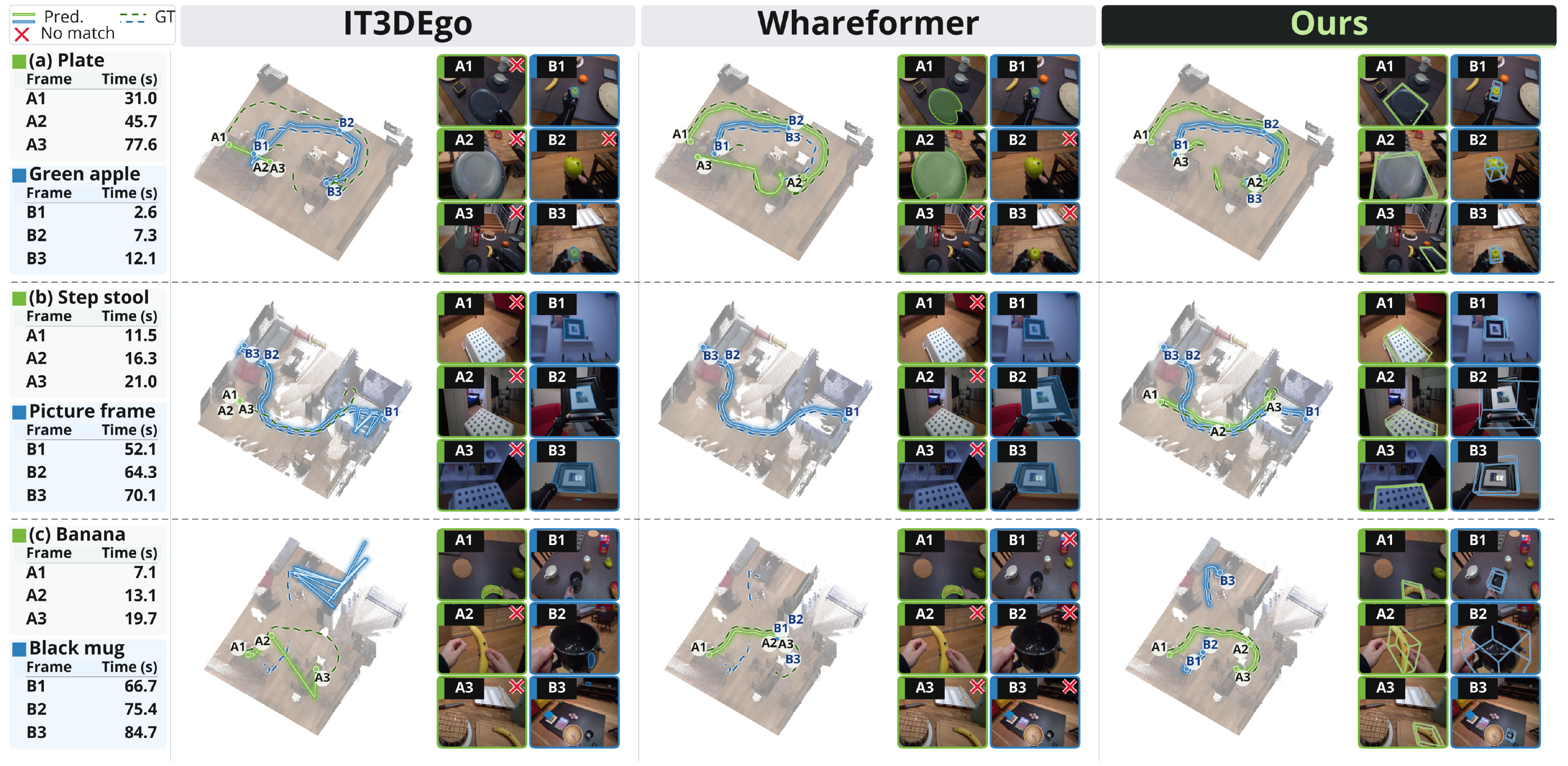}
    \caption{\textbf{Dynamic tracking in three scenes.} Solid/dashed curves
    show predicted/GT trajectories; panels show initial, middle, and final
    observations. Red crosses mark unmatched observations.}
    \label{fig:dynamic_results}
\end{figure}

In Table~\ref{tab:adt_dynamic_tracking}, \textsc{HoliWorld-Foundry} attains HOTA scores of 54.31\% on ADT and 43.14\% on HoliWorld-Bench, compared with 36.00\% and 15.13\% for the strongest baselines, respectively. Higher AssA and PCL indicate more consistent identities and broader trajectory coverage across both datasets. Box metrics provide a complementary comparison of spatial extents on matched objects. Fig.~\ref{fig:dynamic_results} illustrates trajectories that more closely follow the reference paths across the shown interaction stages.

\begin{table}[htbp]
\centering
\caption{\textbf{Dynamic tracking} on ADT20 and HoliWorld-Bench (\%) at a
0.1\,m trajectory tolerance. $^{*}$\textbf{\textit{GT first-visible mask
initialization}}; $^{\dagger}$\textbf{\textit{GT category queries}}. See
Appendix~\ref{app:dynamic_full_metrics} for input adaptations and evaluation
protocols. Best and second-best results are highlighted in
deep green~\rankswatch{rankfirst} and light green~\rankswatch{ranksecond}.}
\label{tab:adt_dynamic_tracking}
\small
\setlength{\tabcolsep}{2pt}
\renewcommand{\arraystretch}{1.15}
\begin{tabular*}{\textwidth}{@{\extracolsep{\fill}}lrrrrrrrrrrrr@{}}
\toprule
& \multicolumn{6}{c}{\textbf{ADT20}} & \multicolumn{6}{c}{\textbf{HoliWorld-Bench}} \\
\cmidrule(lr){2-7}\cmidrule(lr){8-13}
& \multicolumn{3}{c}{Bounding box} & \multicolumn{3}{c}{Trajectory}
& \multicolumn{3}{c}{Bounding box} & \multicolumn{3}{c}{Trajectory} \\
\cmidrule(lr){2-4}\cmidrule(lr){5-7}\cmidrule(lr){8-10}\cmidrule(lr){11-13}
\textbf{Method}
& IoU$\uparrow$ & GTC$\uparrow$ & PP$\uparrow$
& HOTA$\uparrow$ & AssA$\uparrow$ & PCL$\uparrow$
& IoU$\uparrow$ & GTC$\uparrow$ & PP$\uparrow$
& HOTA$\uparrow$ & AssA$\uparrow$ & PCL$\uparrow$ \\
\midrule
\multicolumn{13}{@{}l}{\textit{Instance segmentation and tracking pipeline}} \\
EgoSeg3D$^{\dagger}$ & -- & -- & -- & 21.89 & 37.11 & \second{36.47} & -- & -- & -- & 5.76 & \second{21.78} & \second{36.84} \\
IT3DEgo$^{*}$ & -- & -- & -- & \second{36.00} & \second{53.60} & 34.64 & -- & -- & -- & 2.44 & 4.36 & 7.89 \\
SAM+Whareformer$^{*}$ & -- & -- & -- & 30.76 & 40.99 & 31.26 & -- & -- & -- & 14.63 & 20.97 & \second{36.84} \\
\midrule
\multicolumn{13}{@{}l}{\textit{3D object perception and tracking pipelines}} \\
SAM+LMK$^{*}$ & \second{7.96} & \second{21.27} & \second{14.53} & 26.10 & 30.76 & 21.45 & \second{20.28} & \second{50.84} & \second{21.30} & \second{15.13} & 14.35 & 23.68 \\
\textbf{Ours} & \best{30.88} & \best{43.03} & \best{56.25} & \best{54.31} & \best{71.94} & \best{47.95} & \best{41.79} & \best{53.18} & \best{76.49} & \best{43.14} & \best{56.54} & \best{57.89} \\
\bottomrule
\end{tabular*}
\end{table}

\subsection{Downstream Spatial and Temporal Understanding}
\label{sec:downstream_understanding}

Fine-tuning Qwen3.5-4B~\citep{qwen3.5} on HoliWorld-Data raises Token F1 from 16.35\% to 78.02\% on HoliWorld-Bench and from 18.29\% to 66.39\% on EgoDynamic4D (Table~\ref{tab:spatial_reasoning_results}), supporting the value of the generated supervision for spatial and temporal reasoning. The resulting HoliWorld-VL-4B also surpasses zero-shot Qwen3.8-27B~\citep{qwen38} on all four metrics across both benchmarks. The gains in yes/no, multiple-choice, and numerical accuracy complement the Token F1 improvements, indicating that the generated supervision benefits multiple question formats across the two benchmarks.

\begin{table}[htbp]
\centering
\caption{\textbf{Spatial reasoning results} (\%). F1 denotes Token F1; Y/N,
MCQ, and Num. report yes/no, multiple-choice, and numerical accuracy. Best and
second-best results are highlighted in deep green~\rankswatch{rankfirst} and
light green~\rankswatch{ranksecond}.}
\label{tab:spatial_reasoning_results}
\small
\setlength{\tabcolsep}{4pt}
\renewcommand{\arraystretch}{1.15}
\begin{tabular*}{\textwidth}{@{\extracolsep{\fill}}lcccccccc@{}}
\toprule
& \multicolumn{4}{c}{\textbf{HoliWorld-Bench}} & \multicolumn{4}{c}{\textbf{EgoDynamic4D}} \\
\cmidrule(lr){2-5}\cmidrule(lr){6-9}
\textbf{Model} & F1$\uparrow$ & Y/N$\uparrow$ & MCQ$\uparrow$ & Num.$\uparrow$
& F1$\uparrow$ & Y/N$\uparrow$ & MCQ$\uparrow$ & Num.$\uparrow$ \\
\midrule
Qwen3.5-4B      & 16.35 & 34.86 & 62.00 & 10.80
                & 18.29 & 59.47 & \second{79.16} & 0.02 \\
Qwen3.8-27B     & \second{43.67} & \second{82.85} & \second{75.00} & \second{66.13}
                & \second{42.93} & \second{61.07} & 57.54 & \second{0.16} \\
\textbf{HoliWorld-VL-4B} & \best{78.02} & \best{93.52} & \best{94.00} & \best{81.84}
                & \best{66.39} & \best{99.67} & \best{94.92} & \best{16.33} \\
\bottomrule
\end{tabular*}
\end{table}

\subsection{Ablation Studies}
\label{sec:ablation_study}

\begin{figure}[htbp]
    \centering
    \resizebox{0.90\textwidth}{!}{\input{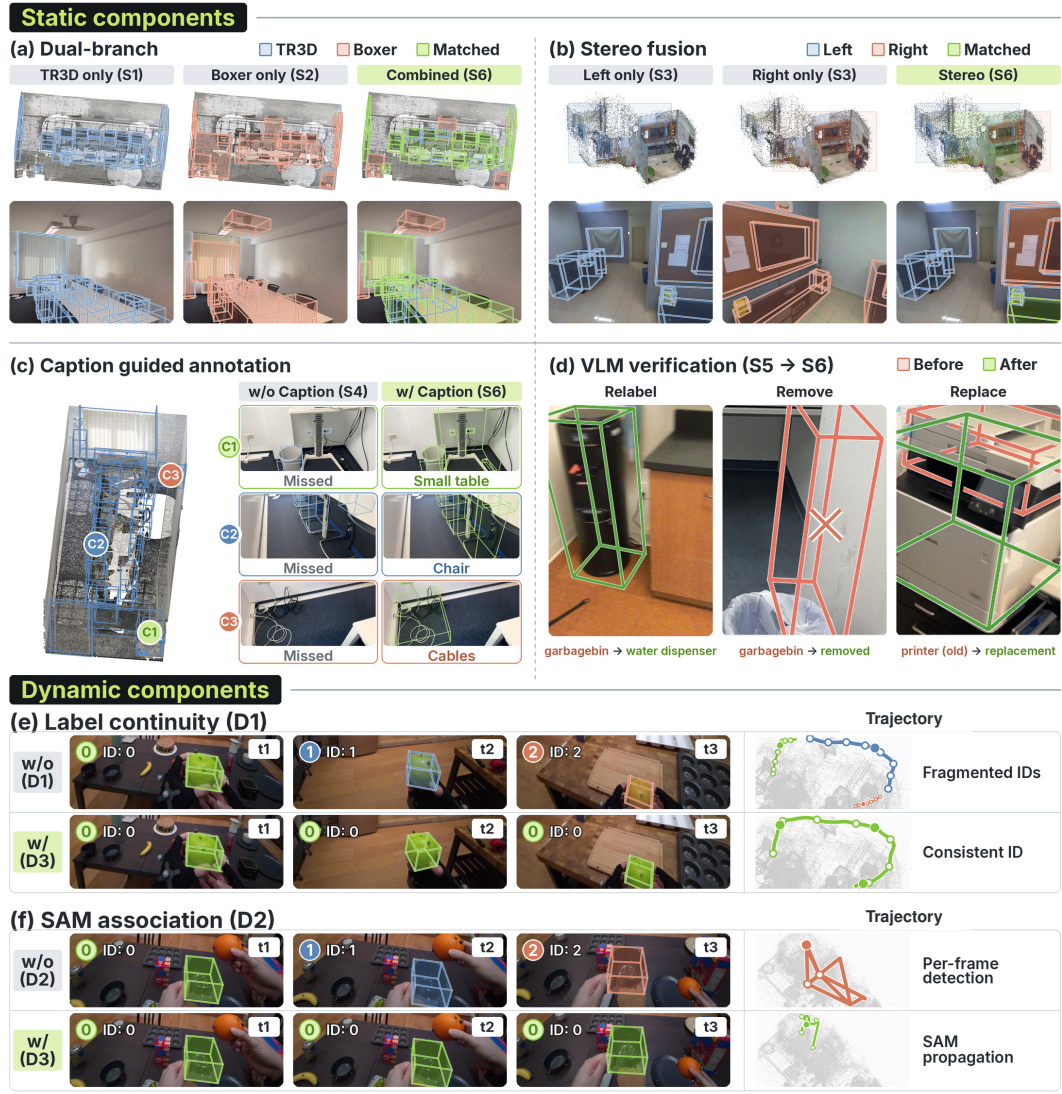}}
    \caption{\textbf{Qualitative ablation results} for the static and dynamic
    components.}
    \label{fig:component_ablation}
\end{figure}

\textbf{Static components.}
Fig.~\ref{fig:component_ablation}(a--b) illustrates complementary geometric coverage: image-based lifting method captures more small objects, point-cloud detection recovers larger extents, and stereo fusion combines observations from both views. Caption guidance recovers missed entities (c). VLM verification shows modest changes in geometric metrics, while the qualitative examples illustrate corrections to object labels and unsuitable candidates (d), suggesting a complementary role in semantic annotation. Detailed AP$_{25}$ and mIoU comparisons are reported in Table~\ref{tab:component_ablation} (Appendix~\ref{app:component_ablation}).

\textbf{Dynamic components.}
Fig.~\ref{fig:component_ablation}(e--f) highlights two complementary requirements for reliable trajectories. Cross-interaction label continuity reduces identity fragmentation across repeated observations, while SAM association supplies temporally consistent masks for 3D lifting, reducing the trajectory deviations seen with per-frame detection. Together, they preserve both object correspondence and spatial continuity. Corresponding HOTA and PCL gains on both datasets are reported in Table~\ref{tab:component_ablation} (Appendix~\ref{app:component_ablation}).

\section{Conclusion}
\label{sec:conclusion}

In this paper, we presented \textsc{HoliWorld}, an integrated ecosystem
that places human egocentric experience at the center of learning world
intelligence. Through complementary static and dynamic perception,
\textsc{HoliWorld}-\textit{Foundry} automatically constructs world records linking
scene geometry, object semantics, and interaction histories to supporting
observations. These records form \textsc{HoliWorld}-\textit{Data} and support
\textsc{HoliWorld}-\textit{VL} training, while independently annotated
\textsc{HoliWorld}-\textit{Bench} provides reference data for evaluating annotation
quality and learned capabilities. Experiments demonstrate improved static
perception and dynamic tracking, alongside gains in spatial and temporal
reasoning on \textsc{HoliWorld}-\textit{Bench} and EgoDynamic4D.
\textsc{HoliWorld} provides a practical foundation for expanding these grounded
learning resources with new recordings, enabling machines to learn how the
world is organized and how it changes through interaction.

\FloatBarrier
{
\small
\bibliography{references}

\begin{thebibliography}{52}
\providecommand{\natexlab}[1]{#1}

\bibitem[{Assran et~al.(2025)Assran, Bardes, Fan et~al.}]{assran2025vjepa2}
Mido Assran, Adrien Bardes, David Fan, and 1 others. 2025.
\newblock \href {https://arxiv.org/abs/2506.09985} {{V-JEPA 2}: Self-supervised
  video models enable understanding, prediction and planning}.
\newblock \emph{arXiv preprint arXiv:2506.09985}.

\bibitem[{Avetisyan et~al.(2025)Avetisyan, Xie, Howard-Jenkins, Yang, Aroudj,
  Patra, Zhang, Frost, Holland, Orme, Engel, Miller, Newcombe, and
  Balntas}]{avetisyan2024scenescript}
Armen Avetisyan, Christopher Xie, Henry Howard-Jenkins, Tsun-Yi Yang, Samir
  Aroudj, Suvam Patra, Fuyang Zhang, Duncan Frost, Luke Holland, Campbell Orme,
  Jakob Engel, Edward Miller, Richard Newcombe, and Vasileios Balntas. 2025.
\newblock \href {https://doi.org/10.1007/978-3-031-73030-6_14} {{SceneScript}:
  Reconstructing scenes with an autoregressive structured language model}.
\newblock In \emph{Computer Vision -- ECCV 2024}, pages 247--263.

\bibitem[{Bai et~al.(2025)Bai, Chen, Liu, Wang, Ge, Song, Dang, Wang, Wang,
  Tang, Zhong, Zhu, Yang, Li, Wan, Wang, Ding, Fu, Xu, Ye, Zhang, Xie, Cheng,
  Zhang, Yang, Xu, and Lin}]{bai2025qwen25vl}
Shuai Bai, Keqin Chen, Xuejing Liu, Jialin Wang, Wenbin Ge, Sibo Song, Kai
  Dang, Peng Wang, Shijie Wang, Jun Tang, Humen Zhong, Yuanzhi Zhu, Mingkun
  Yang, Zhaohai Li, Jianqiang Wan, Pengfei Wang, Wei Ding, Zheren Fu, Yiheng
  Xu, and 8 others. 2025.
\newblock {Qwen2.5-VL} technical report.
\newblock \emph{arXiv preprint arXiv:2502.13923}.

\bibitem[{Bhalgat et~al.(2024)Bhalgat, Tschernezki, Laina, Henriques, Vedaldi,
  and Zisserman}]{bhalgat2024egoseg3d}
Yash Bhalgat, Vadim Tschernezki, Iro Laina, Joao~F. Henriques, Andrea Vedaldi,
  and Andrew Zisserman. 2024.
\newblock {3D}-aware instance segmentation and tracking in egocentric videos.
\newblock In \emph{Asian Conference on Computer Vision}.

\bibitem[{Chalk et~al.(2026)Chalk, Sinha, Damen, Kalantidis, and
  Larlus}]{chalk2026whareformer}
Jacob Chalk, Saptarshi Sinha, Dima Damen, Yannis Kalantidis, and Diane Larlus.
  2026.
\newblock {Whareformer}: Learning to track what is where in long egocentric
  videos.
\newblock In \emph{European Conference on Computer Vision (ECCV)}.

\bibitem[{Chen et~al.(2024)Chen, Xu, Kirmani, Ichter, Sadigh, Guibas, and
  Xia}]{chen2024spatialvlm}
Boyuan Chen, Zhuo Xu, Sean Kirmani, Brian Ichter, Dorsa Sadigh, Leonidas
  Guibas, and Fei Xia. 2024.
\newblock \href
  {https://openaccess.thecvf.com/content/CVPR2024/html/Chen_SpatialVLM_Endowing_Vision-Language_Models_with_Spatial_Reasoning_Capabilities_CVPR_2024_paper.html}
  {{SpatialVLM}: Endowing vision-language models with spatial reasoning
  capabilities}.
\newblock In \emph{Proceedings of the IEEE/CVF Conference on Computer Vision
  and Pattern Recognition (CVPR)}, pages 14455--14465.

\bibitem[{Cheng et~al.(2023)Cheng, Oh, Price, Schwing, and Lee}]{cheng2023deva}
Ho~Kei Cheng, Seoung~Wug Oh, Brian Price, Alexander Schwing, and Joon-Young
  Lee. 2023.
\newblock \href
  {https://openaccess.thecvf.com/content/ICCV2023/html/Cheng_Tracking_Anything_with_Decoupled_Video_Segmentation_ICCV_2023_paper.html}
  {Tracking anything with decoupled video segmentation}.
\newblock In \emph{Proceedings of the IEEE/CVF International Conference on
  Computer Vision (ICCV)}.

\bibitem[{Dai et~al.(2017)Dai, Chang, Savva, Halber, Funkhouser, and
  Nie{\ss}ner}]{scannet}
Angela Dai, Angel~X. Chang, Manolis Savva, Maciej Halber, Thomas Funkhouser,
  and Matthias Nie{\ss}ner. 2017.
\newblock \href
  {https://openaccess.thecvf.com/content_cvpr_2017/html/Dai_ScanNet_Richly-Annotated_3D_CVPR_2017_paper.html}
  {{ScanNet}: Richly-annotated {3D} reconstructions of indoor scenes}.
\newblock In \emph{Proceedings of the IEEE Conference on Computer Vision and
  Pattern Recognition (CVPR)}.

\bibitem[{Damen et~al.(2022)Damen, Doughty, Farinella, Furnari, Kazakos, Ma,
  Moltisanti, Munro, Perrett, Price, and Wray}]{damen2022rescaling}
Dima Damen, Hazel Doughty, Giovanni~Maria Farinella, Antonino Furnari,
  Evangelos Kazakos, Jian Ma, Davide Moltisanti, Jonathan Munro, Toby Perrett,
  Will Price, and Michael Wray. 2022.
\newblock \href {https://doi.org/10.1007/s11263-021-01531-2} {Rescaling
  egocentric vision: Collection, pipeline and challenges for
  {EPIC-KITCHENS-100}}.
\newblock \emph{International Journal of Computer Vision}, 130:33--55.

\bibitem[{Darkhalil et~al.(2022)Darkhalil, Shan, Zhu, Ma, Kar, Higgins, Fidler,
  Fouhey, and Damen}]{darkhalil2022visor}
Ahmad Darkhalil, Dandan Shan, Bin Zhu, Jian Ma, Amlan Kar, Richard Higgins,
  Sanja Fidler, David Fouhey, and Dima Damen. 2022.
\newblock \href {https://arxiv.org/abs/2209.13064} {{EPIC-KITCHENS VISOR}
  benchmark: Video segmentations and object relations}.
\newblock In \emph{Advances in Neural Information Processing Systems},
  volume~35.

\bibitem[{DeTone et~al.(2026)DeTone, Shen, Zhang, Ma, Straub, Newcombe, and
  Engel}]{boxer2026}
Daniel DeTone, Tianwei Shen, Fan Zhang, Lingni Ma, Julian Straub, Richard
  Newcombe, and Jakob Engel. 2026.
\newblock \href {https://github.com/facebookresearch/boxer} {{Boxer}: Robust
  lifting of open-world {2D} bounding boxes to {3D}}.
\newblock In \emph{European Conference on Computer Vision (ECCV)}.

\bibitem[{Gao et~al.(2026)Gao, Li, Liu, Ji, Gong, Liao, Liu, Zhang, Yang, Xu,
  Yang, Huang, Zhang, Liu, Sun, Zhang, and Zhong}]{holispatial2026}
Yuanyuan Gao, Hao Li, Yifei Liu, Xinhao Ji, Yuning Gong, Yuanjun Liao, Fangfu
  Liu, Manyuan Zhang, Yuchen Yang, Dan Xu, Xue Yang, Huaxi Huang, Hongjie
  Zhang, Ziwei Liu, Xiao Sun, Dingwen Zhang, and Zhihang Zhong. 2026.
\newblock \href {https://visionary-laboratory.github.io/holi-spatial/}
  {{Holi-Spatial}: Evolving video streams into holistic {3D} spatial
  intelligence}.
\newblock In \emph{Proceedings of the 43rd International Conference on Machine
  Learning (ICML)}.

\bibitem[{Grauman et~al.(2022)Grauman, Westbury, Byrne, Chavis, Furnari,
  Girdhar, Hamburger, Jiang, Liu, Liu, Martin, Nagarajan, Radosavovic,
  Ramakrishnan, Ryan, Sharma, Wray, Xu, Xu, Zhao, Bansal, Batra, Cartillier,
  Crane, Do, Doulaty, Erapalli, Feichtenhofer, Fragomeni, Fu, Gebreselasie,
  Gonz\'alez, Hillis, Huang, Huang, Jia, Khoo, Kol\'a\v{r}, Kottur, Kumar,
  Landini, Li, Li, Li, Mangalam, Modhugu, Munro, Murrell, Nishiyasu, Price,
  Ruiz, Ramazanova, Sari, Somasundaram, Southerland, Sugano, Tao, Vo, Wang, Wu,
  Yagi, Zhao, Zhu, Arbel\'aez, Crandall, Damen, Farinella, Fuegen, Ghanem,
  Ithapu, Jawahar, Joo, Kitani, Li, Newcombe, Oliva, Park, Rehg, Sato, Shi,
  Shou, Torralba, Torresani, Yan, and Malik}]{grauman2022ego4d}
Kristen Grauman, Andrew Westbury, Eugene Byrne, Zachary Chavis, Antonino
  Furnari, Rohit Girdhar, Jackson Hamburger, Hao Jiang, Miao Liu, Xingyu Liu,
  Miguel Martin, Tushar Nagarajan, Ilija Radosavovic, Santhosh~Kumar
  Ramakrishnan, Fiona Ryan, Jayant Sharma, Michael Wray, Mengmeng Xu,
  Eric~Zhongcong Xu, and 66 others. 2022.
\newblock \href
  {https://openaccess.thecvf.com/content/CVPR2022/html/Grauman_Ego4D_Around_the_World_in_3000_Hours_of_Egocentric_Video_CVPR_2022_paper.html}
  {{Ego4D}: Around the world in 3,000 hours of egocentric video}.
\newblock In \emph{Proceedings of the IEEE/CVF Conference on Computer Vision
  and Pattern Recognition}, pages 18995--19012.

\bibitem[{Grauman et~al.(2024)Grauman, Westbury, Torresani, Kitani, Malik
  et~al.}]{grauman2024egoexo4d}
Kristen Grauman, Andrew Westbury, Lorenzo Torresani, Kris Kitani, Jitendra
  Malik, and 1 others. 2024.
\newblock \href {https://arxiv.org/abs/2311.18259} {{Ego-Exo4D}: Understanding
  skilled human activity from first- and third-person perspectives}.
\newblock In \emph{Proceedings of the IEEE/CVF Conference on Computer Vision
  and Pattern Recognition}.

\bibitem[{Huang et~al.(2026)Huang, Hao, Hu, Wang, and
  Wang}]{huang2026egodynamic4d}
Junsheng Huang, Shengyu Hao, Bo-Cheng Hu, Hongwei Wang, and Gaoang Wang. 2026.
\newblock \href {https://doi.org/10.1609/aaai.v40i7.37416} {Understanding
  dynamic scenes in egocentric {4D} point clouds}.
\newblock \emph{Proceedings of the AAAI Conference on Artificial Intelligence},
  40(7):5031--5039.

\bibitem[{Kerbl et~al.(2023)Kerbl, Kopanas, Leimk{\"u}hler, and
  Drettakis}]{kerbl20233d}
Bernhard Kerbl, Georgios Kopanas, Thomas Leimk{\"u}hler, and George Drettakis.
  2023.
\newblock 3d gaussian splatting for real-time radiance field rendering.
\newblock \emph{ACM Trans. Graph.}, 42(4):139--1.

\bibitem[{Kirillov et~al.(2023)Kirillov, Mintun, Ravi, Mao, Rolland, Gustafson,
  Xiao, Whitehead, Berg, Lo, Doll{\'a}r, and Girshick}]{kirillov2023sam}
Alexander Kirillov, Eric Mintun, Nikhila Ravi, Hanzi Mao, Chloe Rolland, Laura
  Gustafson, Tete Xiao, Spencer Whitehead, Alexander~C. Berg, Wan-Yen Lo, Piotr
  Doll{\'a}r, and Ross Girshick. 2023.
\newblock \href
  {https://openaccess.thecvf.com/content/ICCV2023/html/Kirillov_Segment_Anything_ICCV_2023_paper.html}
  {Segment anything}.
\newblock In \emph{Proceedings of the IEEE/CVF International Conference on
  Computer Vision (ICCV)}, pages 4015--4026.

\bibitem[{Lake et~al.(2017)Lake, Ullman, Tenenbaum, and
  Gershman}]{lake2017machines}
Brenden~M. Lake, Tomer~D. Ullman, Joshua~B. Tenenbaum, and Samuel~J. Gershman.
  2017.
\newblock \href {https://doi.org/10.1017/S0140525X16001837} {Building machines
  that learn and think like people}.
\newblock \emph{Behavioral and Brain Sciences}, 40:e253.

\bibitem[{LeCun(2022)}]{lecun2022path}
Yann LeCun. 2022.
\newblock \href {https://openreview.net/pdf?id=BZ5a1r-kVsf} {A path towards
  autonomous machine intelligence}.
\newblock Position paper, version 0.9.2.

\bibitem[{Li et~al.(2023)Li, Li, Savarese, and Hoi}]{li2023blip2}
Junnan Li, Dongxu Li, Silvio Savarese, and Steven Hoi. 2023.
\newblock {BLIP-2}: Bootstrapping language-image pre-training with frozen image
  encoders and large language models.
\newblock In \emph{Proceedings of the 40th International Conference on Machine
  Learning}, volume 202, pages 19730--19742.

\bibitem[{Lin et~al.(2026)Lin, Chen, Liew, Chen, Li, Zhao, Peng, Guo, Zhou,
  Shi, Feng, and Kang}]{depthanything3}
Haotong Lin, Sili Chen, Jun~Hao Liew, Donny~Y. Chen, Zhenyu Li, Yang Zhao, Sida
  Peng, Hengkai Guo, Xiaowei Zhou, Guang Shi, Jiashi Feng, and Bingyi Kang.
  2026.
\newblock \href
  {https://proceedings.iclr.cc/paper_files/paper/2026/hash/e4cd50120b6d7e8daff1749d6bbaa889-Abstract-Conference.html}
  {{Depth Anything 3}: Recovering the visual space from any views}.
\newblock In \emph{International Conference on Learning Representations
  (ICLR)}.

\bibitem[{Lin et~al.(2025)Lin, Xu, Zhu, Yang, Cao, Ran, Hu, Zhu, Xie, Long, Hu,
  Lin, Wang, and Pang}]{lin2025mmsivideo}
Jingli Lin, Runsen Xu, Shaohao Zhu, Sihan Yang, Peizhou Cao, Yunlong Ran, Miao
  Hu, Chenming Zhu, Yiman Xie, Yilin Long, Wenbo Hu, Dahua Lin, Tai Wang, and
  Jiangmiao Pang. 2025.
\newblock {MMSI-Video-Bench}: A holistic benchmark for video-based spatial
  intelligence.
\newblock \emph{arXiv preprint arXiv:2512.10863}.

\bibitem[{Liu et~al.(2023)Liu, Li, Wu, and Lee}]{liu2023llava}
Haotian Liu, Chunyuan Li, Qingyang Wu, and Yong~Jae Lee. 2023.
\newblock Visual instruction tuning.
\newblock In \emph{Advances in Neural Information Processing Systems},
  volume~36.

\bibitem[{Liu et~al.(2025)Liu, Zeng, Ren, Li, Zhang, Yang, Jiang, Li, Yang, Su,
  Zhu, and Zhang}]{liu2025groundingdino}
Shilong Liu, Zhaoyang Zeng, Tianhe Ren, Feng Li, Hao Zhang, Jie Yang, Qing
  Jiang, Chunyuan Li, Jianwei Yang, Hang Su, Jun Zhu, and Lei Zhang. 2025.
\newblock \href {https://doi.org/10.1007/978-3-031-72970-6_3} {{Grounding
  DINO}: Marrying {DINO} with grounded pre-training for open-set object
  detection}.
\newblock In \emph{Computer Vision -- ECCV 2024}, pages 38--55.

\bibitem[{Luiten et~al.(2021)Luiten, Osep, Dendorfer, Torr, Geiger, Leal-Taixe,
  and Leibe}]{luiten2021hota}
Jonathon Luiten, Aljosa Osep, Patrick Dendorfer, Philip Torr, Andreas Geiger,
  Laura Leal-Taixe, and Bastian Leibe. 2021.
\newblock \href {https://doi.org/10.1007/s11263-020-01375-2} {{HOTA}: A higher
  order metric for evaluating multi-object tracking}.
\newblock \emph{International Journal of Computer Vision}, 129(2):548--578.

\bibitem[{Ma et~al.(2024)Ma, Ye, Hong, Guzov, Jiang, Postyeni, Pesqueira,
  Gamino, Baiyya, Kim, Bailey, Fosas, Liu, Liu, Engel, De~Nardi, and
  Newcombe}]{ma2024nymeria}
Lingni Ma, Yuting Ye, Fangzhou Hong, Vladimir Guzov, Yifeng Jiang, Rowan
  Postyeni, Luis Pesqueira, Alexander Gamino, Vijay Baiyya, Hyo~Jin Kim, Kevin
  Bailey, David~Soriano Fosas, C.~Karen Liu, Ziwei Liu, Jakob Engel, Renzo
  De~Nardi, and Richard Newcombe. 2024.
\newblock \href {https://arxiv.org/abs/2406.09905} {{Nymeria}: A massive
  collection of multimodal egocentric daily motion in the wild}.
\newblock In \emph{European Conference on Computer Vision}.

\bibitem[{Ma et~al.(2025)Ma, Ye, de~Melo, Yuille, and Chen}]{ma2025spatialllm}
Wufei Ma, Luoxin Ye, Celso~M de~Melo, Alan Yuille, and Jieneng Chen. 2025.
\newblock \href
  {https://openaccess.thecvf.com/content/CVPR2025/html/Ma_SpatialLLM_A_Compound_3D-Informed_Design_towards_Spatially-Intelligent_Large_Multimodal_Models_CVPR_2025_paper.html}
  {{SpatialLLM}: A compound {3D}-informed design towards spatially-intelligent
  large multimodal models}.
\newblock In \emph{Proceedings of the IEEE/CVF Conference on Computer Vision
  and Pattern Recognition}, pages 17249--17260.

\bibitem[{Mao et~al.(2025)Mao, Zhong, Fang, Zheng, Tang, Zhu, Tan, and
  Zhou}]{mao2025spatiallm}
Yongsen Mao, Junhao Zhong, Chuan Fang, Jia Zheng, Rui Tang, Hao Zhu, Ping Tan,
  and Zihan Zhou. 2025.
\newblock \href
  {https://proceedings.neurips.cc/paper_files/paper/2025/hash/40223e204a9e18dd8b01d7d11ea97939-Abstract-Conference.html}
  {{SpatialLM}: Training large language models for structured indoor modeling}.
\newblock In \emph{Advances in Neural Information Processing Systems},
  volume~38.

\bibitem[{Pan et~al.(2023)Pan, Charron, Yang, Peters, Whelan, Kong, Parkhi,
  Newcombe, and Ren}]{aria_digital_twin}
Xiaqing Pan, Nicholas Charron, Yongqian Yang, Scott Peters, Thomas Whelan, Chen
  Kong, Omkar Parkhi, Richard Newcombe, and Yuheng~(Carl) Ren. 2023.
\newblock \href
  {https://openaccess.thecvf.com/content/ICCV2023/html/Pan_Aria_Digital_Twin_A_New_Benchmark_Dataset_for_Egocentric_3D_ICCV_2023_paper.html}
  {{Aria Digital Twin}: A new benchmark dataset for egocentric {3D} machine
  perception}.
\newblock In \emph{Proceedings of the IEEE/CVF International Conference on
  Computer Vision (ICCV)}, pages 20133--20143.

\bibitem[{Perrett et~al.(2025)Perrett, Darkhalil, Sinha, Emara, Pollard,
  Parida, Liu, Gatti, Bansal, Flanagan, Chalk, Zhu, Guerrier, Abdelazim, Zhu,
  Moltisanti, Wray, Doughty, and Damen}]{perrett2025hdepic}
Toby Perrett, Ahmad Darkhalil, Saptarshi Sinha, Omar Emara, Sam Pollard,
  Kranti~Kumar Parida, Kaiting Liu, Prajwal Gatti, Siddhant Bansal, Kevin
  Flanagan, Jacob Chalk, Zhifan Zhu, Rhodri Guerrier, Fahd Abdelazim, Bin Zhu,
  Davide Moltisanti, Michael Wray, Hazel Doughty, and Dima Damen. 2025.
\newblock \href
  {https://openaccess.thecvf.com/content/CVPR2025/html/Perrett_HD-EPIC_A_Highly-Detailed_Egocentric_Video_Dataset_CVPR_2025_paper.html}
  {{HD-EPIC}: A highly-detailed egocentric video dataset}.
\newblock In \emph{Proceedings of the IEEE/CVF Conference on Computer Vision
  and Pattern Recognition}, pages 23901--23913.

\bibitem[{Plizzari et~al.(2025)Plizzari, Goel, Perrett, Chalk, Kanazawa, and
  Damen}]{plizzari2025osnom}
Chiara Plizzari, Shubham Goel, Toby Perrett, Jacob Chalk, Angjoo Kanazawa, and
  Dima Damen. 2025.
\newblock Spatial cognition from egocentric video: Out of sight, not out of
  mind.
\newblock In \emph{International Conference on 3D Vision (3DV)}.

\bibitem[{Qin et~al.(2024)Qin, Li, Zhou, Wang, and Pfister}]{qin2024langsplat}
Minghan Qin, Wanhua Li, Jiawei Zhou, Haoqian Wang, and Hanspeter Pfister. 2024.
\newblock {LangSplat}: {3D} language gaussian splatting.
\newblock In \emph{Proceedings of the IEEE/CVF Conference on Computer Vision
  and Pattern Recognition}, pages 20051--20060.

\bibitem[{{Qwen Team}(2026{\natexlab{a}})}]{qwen3.5}
{Qwen Team}. 2026{\natexlab{a}}.
\newblock \href {https://qwen.ai/blog?id=qwen3.5} {{Qwen3.5}: Towards native
  multimodal agents}.

\bibitem[{{Qwen Team}(2026{\natexlab{b}})}]{qwen38}
{Qwen Team}. 2026{\natexlab{b}}.
\newblock \href {https://qwen.ai/blog?id=qwen3.8} {{Qwen3.8-Max}: A new bar for
  coding and cowork}.

\bibitem[{Radford et~al.(2021)Radford, Kim, Hallacy, Ramesh, Goh, Agarwal,
  Sastry, Askell, Mishkin, Clark, Krueger, and Sutskever}]{radford2021clip}
Alec Radford, Jong~Wook Kim, Chris Hallacy, Aditya Ramesh, Gabriel Goh,
  Sandhini Agarwal, Girish Sastry, Amanda Askell, Pamela Mishkin, Jack Clark,
  Gretchen Krueger, and Ilya Sutskever. 2021.
\newblock Learning transferable visual models from natural language
  supervision.
\newblock In \emph{Proceedings of the 38th International Conference on Machine
  Learning}, volume 139, pages 8748--8763.

\bibitem[{Ragusa et~al.(2021)Ragusa, Furnari, Livatino, and
  Farinella}]{ragusa2021meccano}
Francesco Ragusa, Antonino Furnari, Salvatore Livatino, and Giovanni~Maria
  Farinella. 2021.
\newblock \href
  {https://openaccess.thecvf.com/content/WACV2021/html/Ragusa_The_MECCANO_Dataset_Understanding_Human-Object_Interactions_From_Egocentric_Videos_in_WACV_2021_paper.html}
  {The {MECCANO} dataset: Understanding human-object interactions from
  egocentric videos in an industrial-like domain}.
\newblock In \emph{Proceedings of the IEEE/CVF Winter Conference on
  Applications of Computer Vision}, pages 1569--1578.

\bibitem[{Ravi et~al.(2025)Ravi, Gabeur, Hu, Hu, Ryali, Ma, Khedr, R{\"a}dle,
  Rolland, Gustafson, Mintun, Pan, Alwala, Carion, Wu, Girshick, Doll{\'a}r,
  and Feichtenhofer}]{ravi2025sam2}
Nikhila Ravi, Valentin Gabeur, Yuan-Ting Hu, Ronghang Hu, Chaitanya Ryali,
  Tengyu Ma, Haitham Khedr, Roman R{\"a}dle, Chloe Rolland, Laura Gustafson,
  Eric Mintun, Junting Pan, Kalyan~Vasudev Alwala, Nicolas Carion, Chao-Yuan
  Wu, Ross Girshick, Piotr Doll{\'a}r, and Christoph Feichtenhofer. 2025.
\newblock \href
  {https://proceedings.iclr.cc/paper_files/paper/2025/hash/45c1f6a8cbf2da59ebf2c802b4f742cd-Abstract-Conference.html}
  {{SAM 2}: Segment anything in images and videos}.
\newblock In \emph{International Conference on Learning Representations
  (ICLR)}.

\bibitem[{Rukhovich et~al.(2023)Rukhovich, Vorontsova, and
  Konushin}]{rukhovich2023tr3d}
Danila Rukhovich, Anna Vorontsova, and Anton Konushin. 2023.
\newblock Tr3d: Towards real-time indoor 3d object detection.
\newblock In \emph{2023 IEEE International Conference on Image Processing
  (ICIP)}, pages 281--285. IEEE.

\bibitem[{Tenenbaum et~al.(2011)Tenenbaum, Kemp, Griffiths, and
  Goodman}]{tenenbaum2011grow}
Joshua~B. Tenenbaum, Charles Kemp, Thomas~L. Griffiths, and Noah~D. Goodman.
  2011.
\newblock \href {https://doi.org/10.1126/science.1192788} {How to grow a mind:
  Statistics, structure, and abstraction}.
\newblock \emph{Science}, 331(6022):1279--1285.

\bibitem[{Tschernezki et~al.(2023)Tschernezki, Darkhalil, Zhu, Fouhey, Laina,
  Larlus, Damen, and Vedaldi}]{tschernezki2023epicfields}
Vadim Tschernezki, Ahmad Darkhalil, Zhifan Zhu, David Fouhey, Iro Laina, Diane
  Larlus, Dima Damen, and Andrea Vedaldi. 2023.
\newblock \href {https://arxiv.org/abs/2306.08731} {{EPIC Fields}: Marrying
  {3D} geometry and video understanding}.
\newblock In \emph{Advances in Neural Information Processing Systems},
  volume~36.

\bibitem[{Wang et~al.(2026)Wang, Yin, Zhang, Zhang, Wang, Wang, Zhang,
  Chandrasegaran, Liu, Krishna, Xie, Wu, Li, and Li}]{wang2026mindcube}
Qineng Wang, Baiqiao Yin, Pingyue Zhang, Jianshu Zhang, Kangrui Wang, Zihan
  Wang, Jieyu Zhang, Keshigeyan Chandrasegaran, Han Liu, Ranjay Krishna,
  Saining Xie, Jiajun Wu, Fei-Fei Li, and Manling Li. 2026.
\newblock Spatial mental modeling from limited views.
\newblock In \emph{International Conference on Learning Representations}.

\bibitem[{Wu et~al.(2025)Wu, Liu, Hung, and Duan}]{wu2025spatialmllm}
Diankun Wu, Fangfu Liu, Yi-Hsin Hung, and Yueqi Duan. 2025.
\newblock \href {https://doi.org/10.52202/085713-0457} {{Spatial-MLLM}:
  Boosting {MLLM} capabilities in visual-based spatial intelligence}.
\newblock In \emph{Advances in Neural Information Processing Systems},
  volume~38.

\bibitem[{Wu et~al.(2026)Wu, Liu, Liu, and Shen}]{wu2026stthink}
Peiran Wu, Yunze Liu, Miao Liu, and Junxiao Shen. 2026.
\newblock \href
  {https://openaccess.thecvf.com/content/WACV2026/html/Wu_ST-Think_How_Multimodal_Large_Language_Models_Reason_About_4D_Worlds_WACV_2026_paper.html}
  {{ST-Think}: How multimodal large language models reason about {4D} worlds
  from ego-centric videos}.
\newblock In \emph{Proceedings of the IEEE/CVF Winter Conference on
  Applications of Computer Vision}, pages 5174--5183.

\bibitem[{Yang et~al.(2025{\natexlab{a}})Yang, Yang, Gupta, Han, Li, and
  Xie}]{yang2025vsi}
Jihan Yang, Shusheng Yang, Anjali~W. Gupta, Rilyn Han, Fei-Fei Li, and Saining
  Xie. 2025{\natexlab{a}}.
\newblock Thinking in space: How multimodal large language models see,
  remember, and recall spaces.
\newblock In \emph{Proceedings of the IEEE/CVF Conference on Computer Vision
  and Pattern Recognition}, pages 10632--10643.

\bibitem[{Yang et~al.(2025{\natexlab{b}})Yang, Liu, Guo, Dong, Zhang, Zhang,
  Wang, Zhou, Xie, Wang, Ouyang, Lin, Cominelli, Cai, Li, Zhang, Zhang, Hong,
  Widmer, Gringoli, Yang, and Liu}]{yang2025egolife}
Jingkang Yang, Shuai Liu, Hongming Guo, Yuhao Dong, Xiamengwei Zhang, Sicheng
  Zhang, Pengyun Wang, Zitang Zhou, Binzhu Xie, Ziyue Wang, Bei Ouyang, Zhengyu
  Lin, Marco Cominelli, Zhongang Cai, Bo~Li, Yuanhan Zhang, Peiyuan Zhang,
  Fangzhou Hong, Joerg Widmer, and 3 others. 2025{\natexlab{b}}.
\newblock \href
  {https://openaccess.thecvf.com/content/CVPR2025/html/Yang_EgoLife_Towards_Egocentric_Life_Assistant_CVPR_2025_paper.html}
  {{EgoLife}: Towards egocentric life assistant}.
\newblock In \emph{Proceedings of the IEEE/CVF Conference on Computer Vision
  and Pattern Recognition}, pages 28885--28900.

\bibitem[{Yang et~al.(2026)Yang, Piccinelli, Rota~Bul\`o, Hong, Rozumny,
  Sch\"onberger, Bauer, Blum, Kontschieder, and Pollefeys}]{yang2026mapdet3d}
Yung-Hsu Yang, Luigi Piccinelli, Samuel Rota~Bul\`o, Sunghwan Hong, Denis
  Rozumny, Johannes Sch\"onberger, Zuria Bauer, Hermann Blum, Peter
  Kontschieder, and Marc Pollefeys. 2026.
\newblock \href {https://royyang0714.github.io/Map-Det3D/} {{Map-Det3D}: Metric
  feed-forward {3D} reconstruction prior for multi-view {3D} object detection
  from streaming inputs}.
\newblock In \emph{European Conference on Computer Vision (ECCV)}.

\bibitem[{Yeshwanth et~al.(2023)Yeshwanth, Liu, Nie{\ss}ner, and
  Dai}]{scannetpp}
Chandan Yeshwanth, Yueh-Cheng Liu, Matthias Nie{\ss}ner, and Angela Dai. 2023.
\newblock \href {https://scannetpp.mlsg.cit.tum.de/scannetpp/} {{ScanNet++}: A
  high-fidelity dataset of {3D} indoor scenes}.
\newblock In \emph{Proceedings of the International Conference on Computer
  Vision (ICCV)}.

\bibitem[{Zhao et~al.(2024)Zhao, Ma, Kong, and Fowlkes}]{zhao2024it3dego}
Yunhan Zhao, Haoyu Ma, Shu Kong, and Charless Fowlkes. 2024.
\newblock Instance tracking in {3D} scenes from egocentric videos.
\newblock In \emph{Proceedings of the IEEE/CVF Conference on Computer Vision
  and Pattern Recognition}.

\bibitem[{Zheng et~al.(2025)Zheng, Huang, and Wang}]{zheng2025video3dllm}
Duo Zheng, Shijia Huang, and Liwei Wang. 2025.
\newblock \href
  {https://openaccess.thecvf.com/content/CVPR2025/html/Zheng_Video-3D_LLM_Learning_Position-Aware_Video_Representation_for_3D_Scene_Understanding_CVPR_2025_paper.html}
  {{Video-3D LLM}: Learning position-aware video representation for {3D} scene
  understanding}.
\newblock In \emph{Proceedings of the IEEE/CVF Conference on Computer Vision
  and Pattern Recognition}, pages 8995--9006.

\bibitem[{Zhu et~al.(2025)Zhu, Wang, Zhang, Pang, and Liu}]{zhu2024llava3d}
Chenming Zhu, Tai Wang, Wenwei Zhang, Jiangmiao Pang, and Xihui Liu. 2025.
\newblock \href
  {https://openaccess.thecvf.com/content/ICCV2025/html/Zhu_LLaVA-3D_A_Simple_yet_Effective_Pathway_to_Empowering_LMMs_with_ICCV_2025_paper.html}
  {{LLaVA-3D}: A simple yet effective pathway to empowering {LMMs} with {3D}
  capabilities}.
\newblock In \emph{Proceedings of the IEEE/CVF International Conference on
  Computer Vision (ICCV)}, pages 4295--4305.

\bibitem[{Zhu et~al.(2026)Zhu, Xi, Ni, Cai, Gong, Zhao, Qu, Miao, Li, Zhong,
  Jiang, and Patel}]{zhu2026egoreasoner}
Fangrui Zhu, Yunfeng Xi, Jianmo Ni, Mu~Cai, Boqing Gong, Long Zhao, Chen Qu,
  Ian Miao, Yi~Li, Cheng Zhong, Huaizu Jiang, and Shwetak Patel. 2026.
\newblock \href {https://arxiv.org/abs/2603.06561} {{EgoReasoner}: Learning
  egocentric {4D} reasoning via task-adaptive structured thinking}.
\newblock \emph{arXiv preprint arXiv:2603.06561}.

\bibitem[{Zou et~al.(2025)Zou, Song, Qiu, Peng, Ye, Liu, and Wang}]{zou2025m3}
Xueyan Zou, Yuchen Song, Ri-Zhao Qiu, Xuanbin Peng, Jianglong Ye, Sifei Liu,
  and Xiaolong Wang. 2025.
\newblock \href
  {https://proceedings.iclr.cc/paper_files/paper/2025/hash/639f9db06bf70c4ff44cf22a7f92cb08-Abstract-Conference.html}
  {{3D}-spatial multimodal memory}.
\newblock In \emph{International Conference on Learning Representations
  (ICLR)}.

\end{thebibliography}
}

\clearpage
\appendix
  \startcontents[appendix]
  \printcontents[appendix]{l}{1}{%
    \section*{Appendix Contents}%
    \vspace{-0.5em}%
  }
  \clearpage

\section{Experimental Details}
\label{app:experimental_details}

\subsection{Static Perception Evaluation Details}
\label{app:static_evaluation}

\textbf{Data and metrics.}
We evaluate on 20 sequences from each of ScanNet and ScanNet++, using their official GT 3D annotations, and 20 self-collected in-house sequences with manually annotated GT. AP$_{25}$ and AP$_{50}$ use 3D IoU thresholds of 0.25 and 0.50, respectively. Matched-box mIoU is averaged over box pairs matched at IoU 0.25. All static metrics use a $[0,1]$ scale; dashes indicate unavailable results.

\textbf{Baselines.}
The comparison covers model-based methods, including the generalist scene models SpatialLM~\citep{mao2025spatiallm}, LLaVA-3D~\citep{zhu2024llava3d}, and SceneScript~\citep{avetisyan2024scenescript}, and the specialist detector Map-Det3D~\citep{yang2026mapdet3d}; 3DGS-based methods M3-Spatial~\citep{zou2025m3} and LangSplat~\citep{qin2024langsplat}, which encode semantics in Gaussian representations; and spatial pipelines Boxer~\citep{boxer2026} and Holi-Spatial~\citep{holispatial2026}, which integrate multiple perception and geometric processing stages.

\begin{figure}[p]
    \centering
    \includegraphics[width=\textwidth,height=0.44\textheight,keepaspectratio]{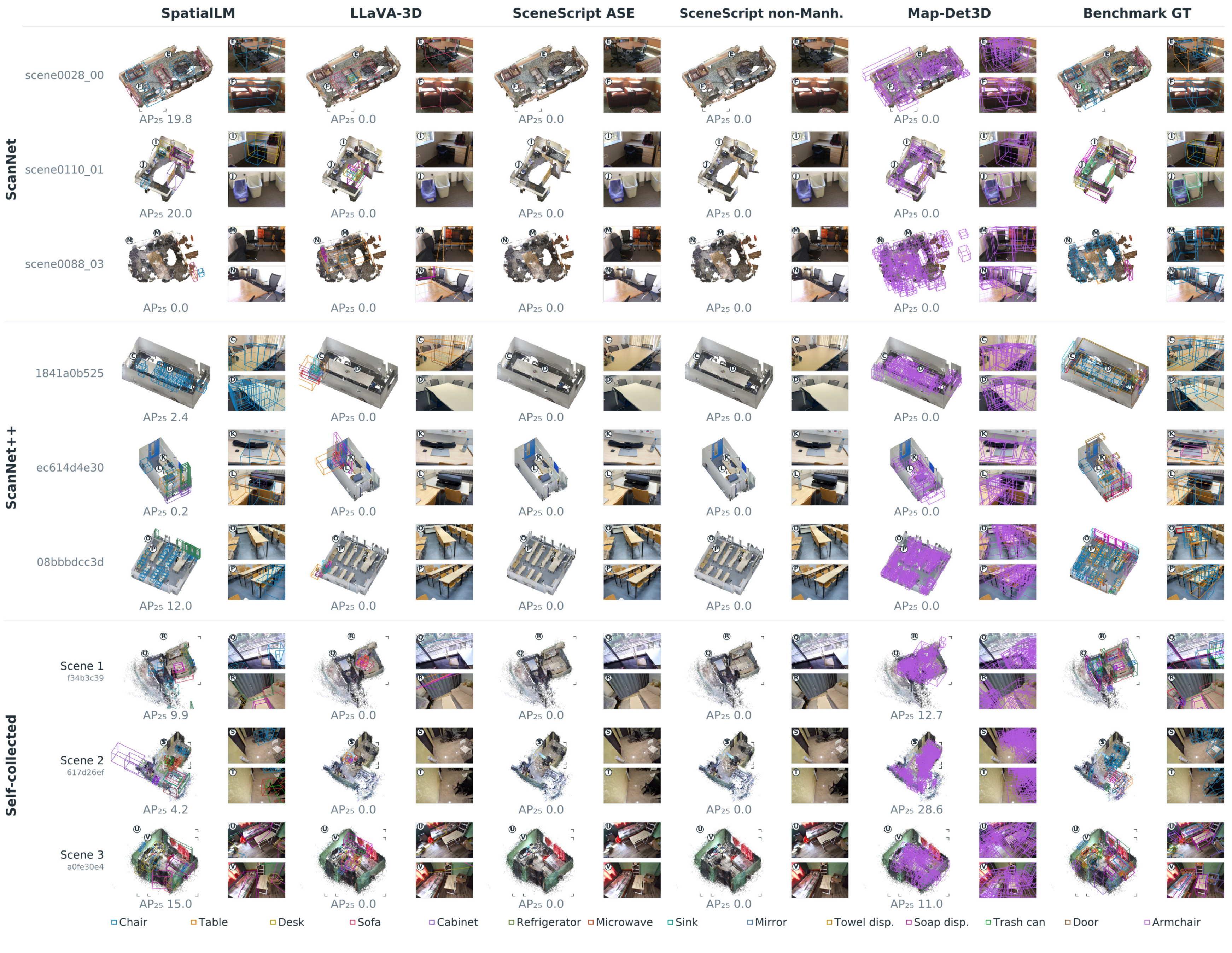}\par
    {\small (a)}\par\vspace{1.5ex}
    \includegraphics[width=\textwidth,height=0.44\textheight,keepaspectratio]{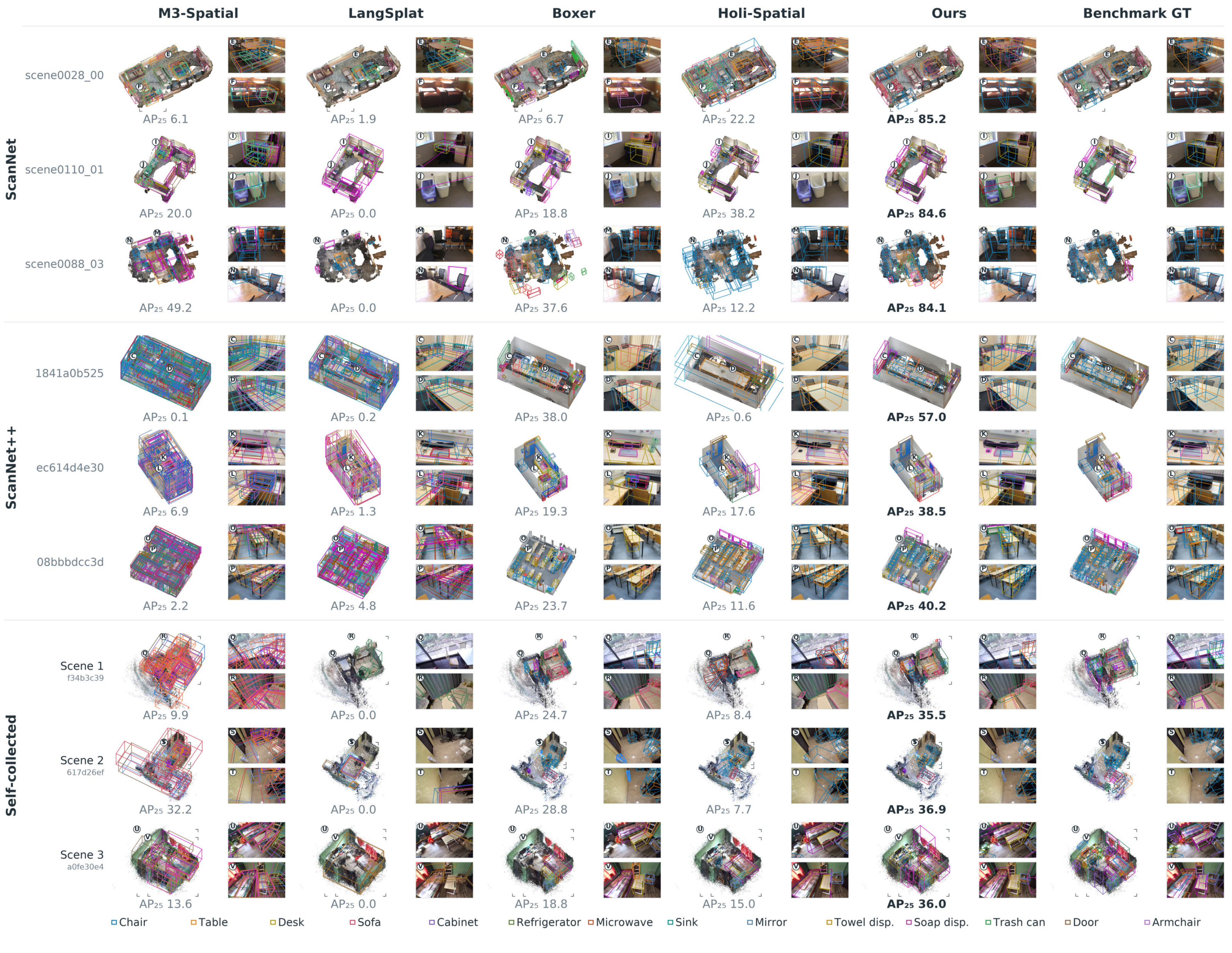}\par
    {\small (b)}
    \caption{\textbf{Qualitative comparison of static 3D object perception}
    across methods and scenes. Panels (a) and (b) show complementary subsets of
    the evaluated methods.}
    \label{fig:static_full_methods}
\end{figure}

\textbf{Additional analysis.}
At IoU 0.50, \textsc{HoliWorld-Foundry} reaches AP$_{50}$ of 0.55 on ScanNet and 0.13 on ScanNet++, compared with the strongest baseline scores of 0.12 and 0.05 (Table~\ref{tab:static_3d_bbox}). These gains persist under stricter geometric matching. On the in-house data, the improvement over Boxer is more pronounced in AP than in matched-box mIoU, suggesting that improved detection contributes beyond the extent accuracy of successfully matched boxes. Because mIoU is conditional on matching, it should be read jointly with AP when assessing overall perception quality.
Fig.~\ref{fig:static_full_methods} compares all evaluated methods qualitatively.

\subsection{Full Dynamic Tracking Results}
\label{app:dynamic_full_metrics}

\begin{figure}[htbp]
    \centering
    \includegraphics[width=\textwidth]{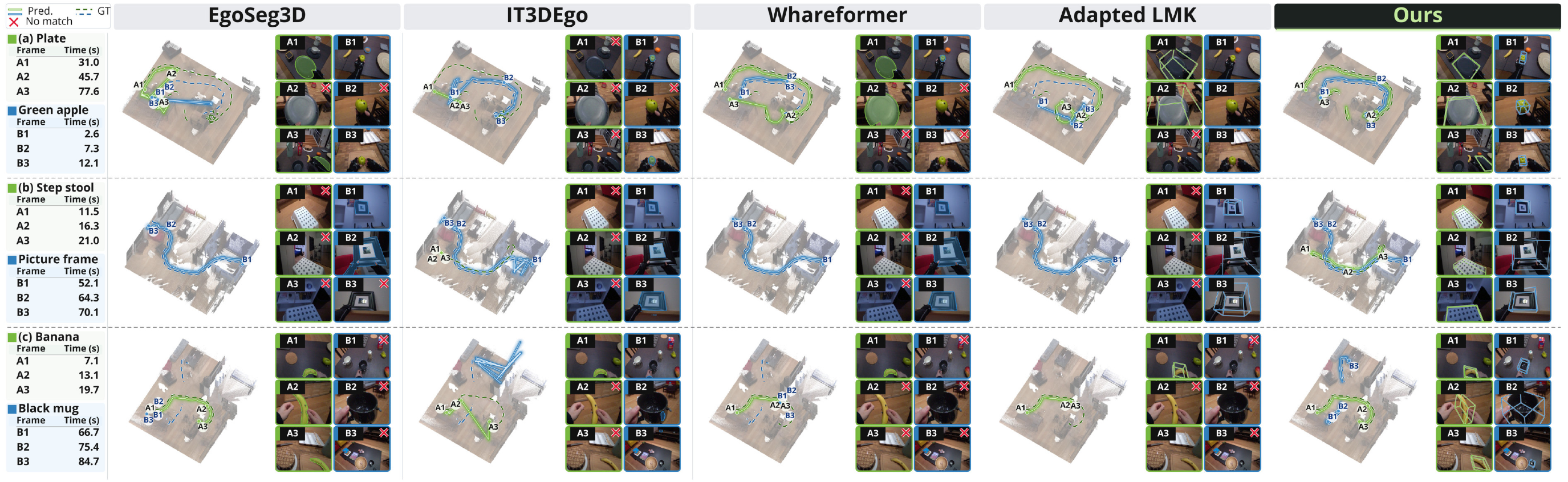}
    \caption{\textbf{Qualitative dynamic tracking comparison} across all
    evaluated methods.}
    \label{fig:dynamic_full_methods}
\end{figure}

Table~\ref{tab:dynamic_full_metrics} and Figure~\ref{fig:dynamic_full_methods} report the full quantitative and qualitative comparisons.

\textbf{Metrics.}
HOTA$_{0.1}$, DetA$_{0.1}$, and AssA$_{0.1}$ follow \citet{luiten2021hota}, using a 0.1\,m world-coordinate center-distance tolerance. They measure joint detection and association quality, detection accuracy, and identity association consistency, respectively. PCL$_{0.1}$ measures correctly localized GT instance-frames under a fixed, sequence-global one-to-one GT--prediction assignment that maximizes matches within 0.1\,m; unassigned trajectories, missing predictions, and localization failures count as incorrect. Recall R$_{0.1}$, precision P$_{0.1}$, and TP-L2 use HOTA's frame-level matches to measure GT coverage, prediction reliability, and mean center error among successful matches. Box geometry is assessed separately using IoU, GT coverage (GTC), and prediction precision (PP). All metrics are percentages except TP-L2, which is in meters.

\textbf{Baselines and inputs.}
EgoSeg3D$^{\dagger}$~\citep{bhalgat2024egoseg3d} receives sequence-level GT category queries. Its GroundingDINO--SAM-v1--DEVA frontend~\citep{liu2025groundingdino,kirillov2023sam,cheng2023deva} generates and propagates masks before 3D lifting and native association. For IT3DEgo$^{*}$~\citep{zhao2024it3dego}, we provide a predefined target set and convert each target's first-visible GT mask to a bounding box for its native SVOE enrollment interface; subsequent mask/proposal inference and memory updates use the original pipeline.

SAM+Whareformer$^{*}$~\citep{chalk2026whareformer} and SAM+LMK$^{*}$~\citep{plizzari2025osnom} share a SAM2 propagation frontend~\citep{ravi2025sam2}, initialized with each target's GT mask at its first visible frame in the video, irrespective of when it starts moving. Propagation continues through the remaining video to supply mask observations for native association. An adapter derives Whareformer's world centers from masks, depth, and camera poses; LMK retains its native 3D center/scale outputs. Ours uses no annotation-derived test-time inputs. GT initialization defines baseline inputs only, not evaluation correspondences.

\textbf{Evaluation and binding rules.}
We evaluate 20 sequences each from ADT and HoliWorld-Bench over the full RGB timeline, including out-of-view periods, wherever a valid GT pose is available within 20\,ms. Missing predictions remain evaluation failures, and sequence scores are averaged equally. We group pipelines by whether they output 3D boxes. Output-dependent GT--prediction binding uses nearest world-coordinate centers; for pipelines with 3D boxes, candidate tracks must additionally have maximum 3D IoU $>0.5$. IoU, GTC, and PP are computed on the resulting bound-object subset. This binding does not filter trajectory evaluation: all methods use the common distance-based scoring described above, with native predicted identities and PCL's own sequence-global assignment.

\begin{table}[htbp]
\centering
\caption{\textbf{Full dynamic tracking results} on ADT20 and HoliWorld-Bench
(\%; TP-L2 in meters). $^{*}$\textbf{\textit{GT first-visible mask
initialization}} (converted to boxes for IT3DEgo);
$^{\dagger}$\textbf{\textit{GT category queries}}. Ours uses no
annotation-derived test-time inputs. Box metrics use bound objects; dashes
denote unavailable results. Best and second-best results are highlighted in
deep green~\rankswatch{rankfirst} and light green~\rankswatch{ranksecond}.}
\label{tab:dynamic_full_metrics}
\small
\setlength{\tabcolsep}{2pt}
\renewcommand{\arraystretch}{1.15}
\begin{tabular*}{\textwidth}{@{\extracolsep{\fill}}lrrrrrrrrrr@{}}
\toprule
& \multicolumn{3}{c}{\textbf{Bounding box}} & \multicolumn{7}{c}{\textbf{Trajectory}} \\
\cmidrule(lr){2-4}\cmidrule(lr){5-11}
\textbf{Method} & IoU$\uparrow$ & GTC$\uparrow$ & PP$\uparrow$
& HOTA$_{0.1}\uparrow$ & DetA$_{0.1}\uparrow$ & AssA$_{0.1}\uparrow$ & PCL$_{0.1}\uparrow$
& R$_{0.1}\uparrow$ & P$_{0.1}\uparrow$ & TP-L2$\downarrow$ \\
\midrule
\multicolumn{11}{@{}l}{\textbf{ADT20}} \\
\multicolumn{11}{@{}l}{\textit{2D instance tracking pipelines}} \\
EgoSeg3D$^{\dagger}$ & -- & -- & -- & 21.89 & 13.78 & 37.11 & \second{36.47} & \second{46.96} & 16.67 & \second{0.0432} \\
IT3DEgo$^{*}$ & -- & -- & -- & \second{36.00} & \second{24.63} & \second{53.60} & 34.64 & 35.69 & 40.93 & 0.0490 \\
SAM+Whareformer$^{*}$ & -- & -- & -- & 30.76 & 23.60 & 40.99 & 31.26 & 33.35 & 41.33 & 0.0497 \\
\multicolumn{11}{@{}l}{\textit{3D object perception and tracking pipelines}} \\
SAM+LMK$^{*}$ & \second{7.96} & \second{21.27} & \second{14.53} & 26.10 & 22.84 & 30.76 & 21.45 & 27.87 & \second{52.17} & 0.0485 \\
\textbf{Ours} & \best{30.88} & \best{43.03} & \best{56.25} & \best{54.31} & \best{42.13} & \best{71.94} & \best{47.95} & \best{49.56} & \best{76.68} & \best{0.0333} \\
\midrule
\multicolumn{11}{@{}l}{\textbf{HoliWorld-Bench (in-house)}} \\
\multicolumn{11}{@{}l}{\textit{2D instance tracking pipelines}} \\
EgoSeg3D$^{\dagger}$ & -- & -- & -- & 5.76 & 1.52 & \second{21.78} & \second{36.84} & \second{42.11} & 1.55 & \best{0.0094} \\
IT3DEgo$^{*}$ & -- & -- & -- & 2.44 & 1.36 & 4.36 & 7.89 & 7.89 & 1.62 & 0.0479 \\
SAM+Whareformer$^{*}$ & -- & -- & -- & 14.63 & 10.20 & 20.97 & \second{36.84} & 39.47 & 12.10 & \second{0.0113} \\
\multicolumn{11}{@{}l}{\textit{3D object perception and tracking pipelines}} \\
SAM+LMK$^{*}$ & \second{20.28} & \second{50.84} & \second{21.30} & \second{15.13} & \second{15.96} & 14.35 & 23.68 & 39.47 & \second{21.13} & \second{0.0113} \\
\textbf{Ours} & \best{41.79} & \best{53.18} & \best{76.49} & \best{43.14} & \best{32.91} & \best{56.54} & \best{57.89} & \best{68.42} & \best{38.81} & 0.0280 \\
\bottomrule
\end{tabular*}
\end{table}

\subsection{Additional Ablation Results}
\label{app:static_ablation}
\label{app:component_ablation}

\begin{figure}[htbp]
    \centering
    \includegraphics[width=\textwidth]{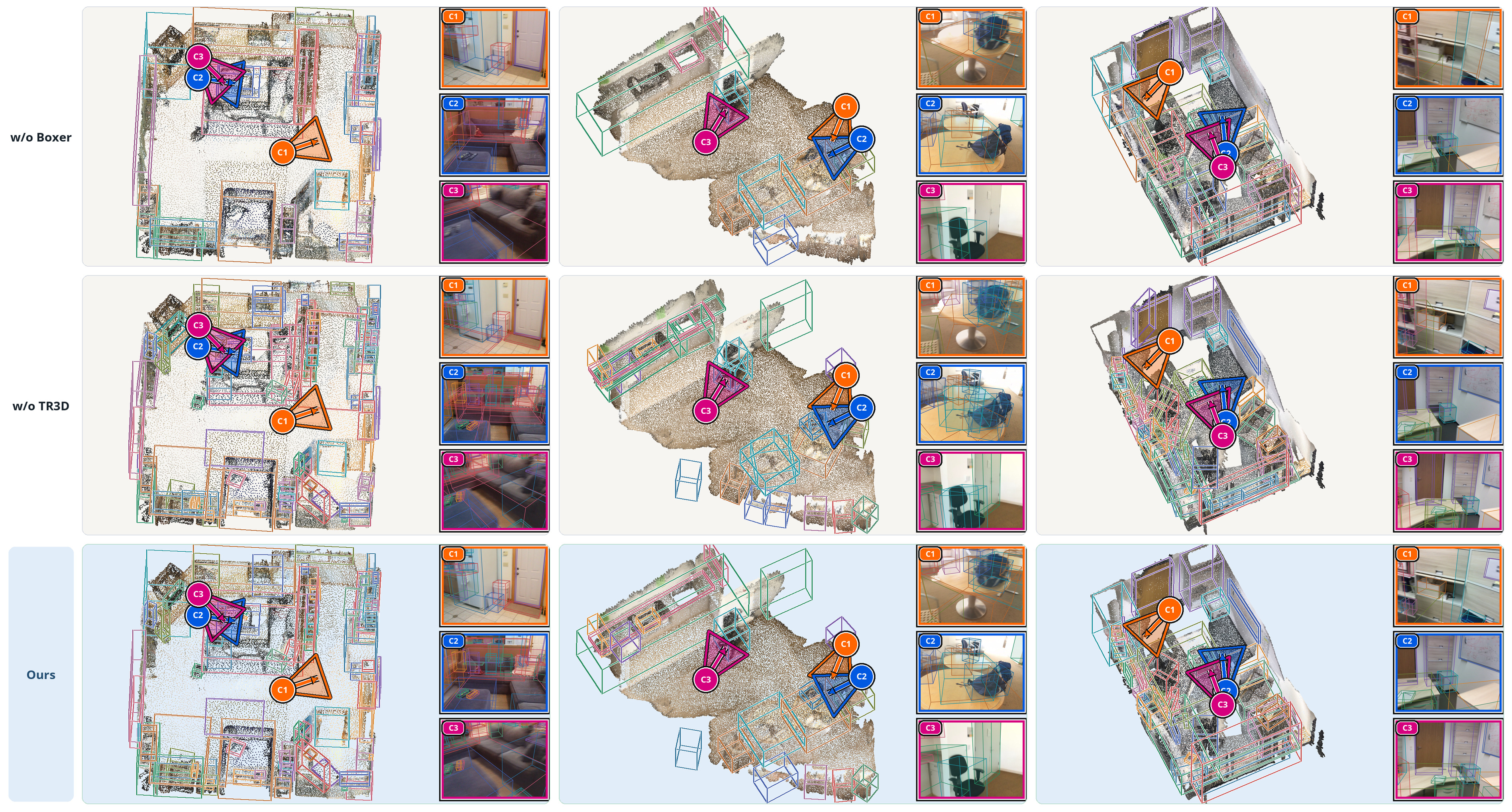}
    \caption{\textbf{Qualitative ablation of dual-branch static object
    perception} across three scenes. Rows show results without 3D lifting (w/o
    Boxer), without point-cloud detection (w/o TR3D), and with both branches
    (Ours). C1--C3 identify camera viewpoints and their corresponding image
    projections.}
    \label{fig:static_ablation}
\end{figure}

Table~\ref{tab:component_ablation} reports component-wise results; Figure~\ref{fig:static_ablation} compares the two static perception branches.

\begin{table}[htbp]
\centering
\caption{\textbf{Component ablations.} Static AP$_{25}$/mIoU: $[0,1]$;
dynamic HOTA/PCL: \% at 0.1\,m. Checkmarks/crosses: enabled/disabled; dashes:
unavailable or inapplicable. Shaded rows (S6, D3) enable all components.}
\label{tab:component_ablation}
\small
\setlength{\tabcolsep}{3.6pt}
\renewcommand{\arraystretch}{1.18}
\newcolumntype{M}{>{\centering\arraybackslash}p{1.08cm}}
\begin{tabular}{@{}cccccccc@{\hspace{6pt}}MMMMMMMM@{}}
\toprule
& \multicolumn{7}{c}{\textbf{Components}}
& \multicolumn{4}{c}{\textbf{Static data}}
& \multicolumn{4}{c}{\textbf{Dynamic data}} \\
\cmidrule(lr){2-8}
\cmidrule(lr){9-12}\cmidrule(lr){13-16}
& \multicolumn{5}{c}{Static} & \multicolumn{2}{c}{Dynamic}
& \multicolumn{2}{c}{ScanNet} & \multicolumn{2}{c}{HoliWorld-Bench}
& \multicolumn{2}{c}{ADT20} & \multicolumn{2}{c}{HoliWorld-Bench} \\
\cmidrule(lr){2-6}\cmidrule(lr){7-8}
\cmidrule(lr){9-10}\cmidrule(lr){11-12}
\cmidrule(lr){13-14}\cmidrule(lr){15-16}
ID & \rotatebox{90}{Boxer} & \rotatebox{90}{TR3D} & \rotatebox{90}{StereoGeo}
& \rotatebox{90}{Caption} & \rotatebox{90}{VLM-Ver.}
& \rotatebox{90}{Label-Cont.} & \rotatebox{90}{SAM-Assoc.}
& AP$_{25}\uparrow$ & mIoU$\uparrow$ & AP$_{25}\uparrow$ & mIoU$\uparrow$
& HOTA$\uparrow$ & PCL$\uparrow$ & HOTA$\uparrow$ & PCL$\uparrow$ \\
\midrule
S1 & \xmark & \cmark & \cmark & \cmark & \cmark & -- & --
   & 0.58 & 0.80 & 0.10 & 0.47 & -- & -- & -- & -- \\
S2 & \cmark & \xmark & \cmark & \cmark & \cmark & -- & --
   & 0.32 & 0.30 & 0.20 & 0.52 & -- & -- & -- & -- \\
S3 & \cmark & \cmark & \xmark & \cmark & \cmark & -- & --
   & -- & -- & 0.20 & 0.53 & -- & -- & -- & -- \\
S4 & \cmark & \cmark & \cmark & \xmark & \cmark & -- & --
   & 0.67 & 0.70 & 0.21 & 0.52 & -- & -- & -- & -- \\
S5 & \cmark & \cmark & \cmark & \cmark & \xmark & -- & --
   & 0.68 & 0.72 & 0.24 & 0.52 & -- & -- & -- & -- \\
\rowcolor{fullconfig}
S6 & \cmark & \cmark & \cmark & \cmark & \cmark & -- & --
   & 0.68 & 0.72 & 0.24 & 0.53 & -- & -- & -- & -- \\
\midrule
D1 & -- & -- & -- & -- & -- & \xmark & \cmark
   & -- & -- & -- & -- & 44.09 & 42.62 & 32.86 & 45.30 \\
D2 & -- & -- & -- & -- & -- & \cmark & \xmark
   & -- & -- & -- & -- & 38.38 & 34.08 & 38.89 & 51.04 \\
\rowcolor{fullconfig}
D3 & -- & -- & -- & -- & -- & \cmark & \cmark
   & -- & -- & -- & -- & 54.31 & 47.95 & 43.14 & 57.89 \\
\bottomrule
\end{tabular}
\end{table}

\subsection{Holi-World Static and Dynamic QA}
\label{app:qa_types}

\begin{figure}[htbp]
    \centering
    \includegraphics[width=\textwidth]{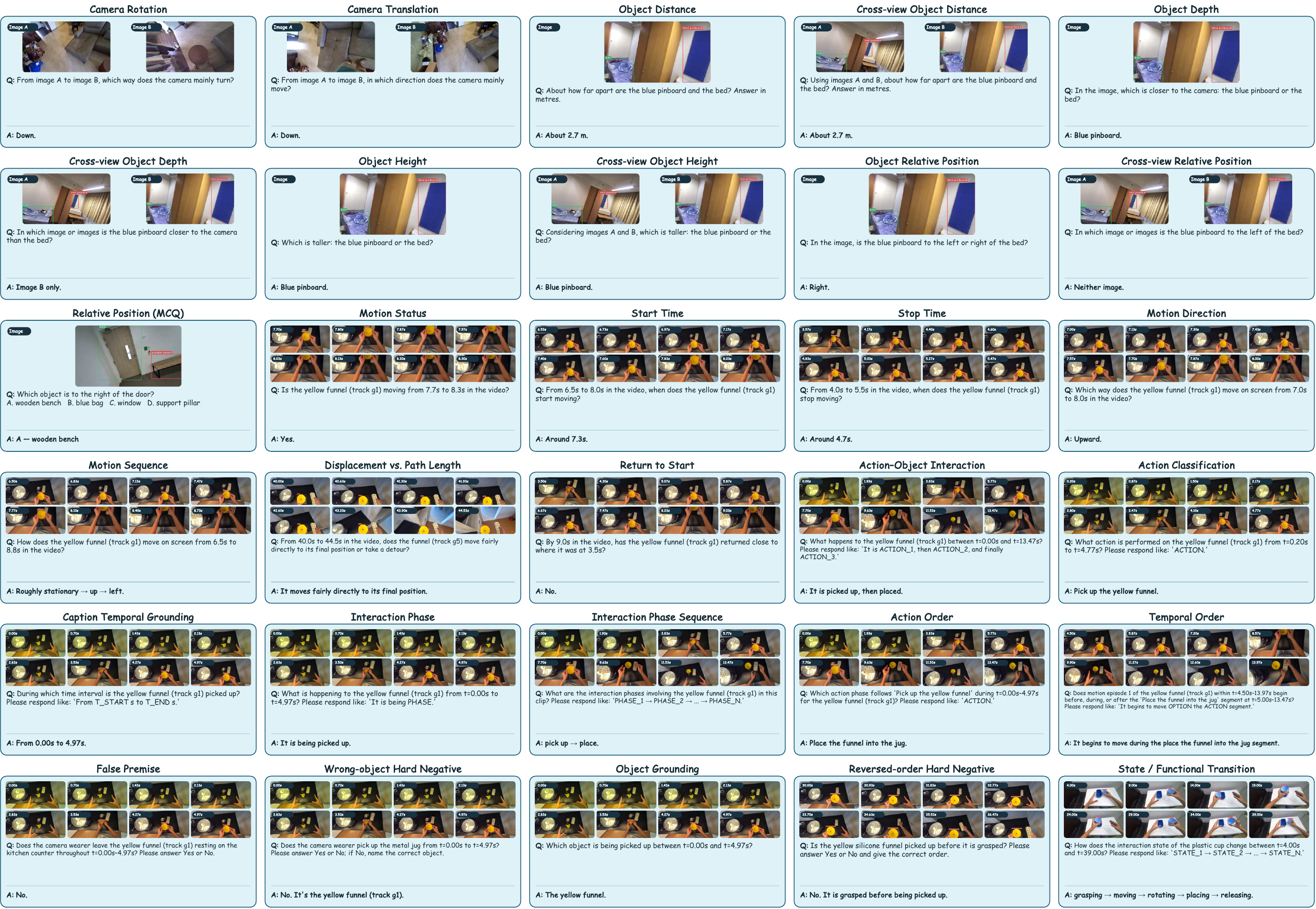}
    \caption{\textbf{Examples of the 30 QA types in Holi-World}, covering
    static mapping, dynamic trajectory, and dynamic caption reasoning.}
    \label{fig:qa_examples}
\end{figure}

\textbf{30 QA types.}
We design 30 fine-grained QA types, broadly divided into static and dynamic reasoning. The static category contains 11 types that evaluate camera motion and object-level spatial relationships, including distance, depth, height, and relative position under single- and multi-view settings. The dynamic category comprises seven trajectory-based types for motion status, timing, direction, sequence, and path characteristics, together with 12 caption-based types covering action recognition, object interaction, temporal grounding, interaction phases, event ordering, state transitions, and contrastive hard negatives. As illustrated in Fig.~\ref{fig:qa_examples}, these questions combine visual observations with 3D bounding boxes, object trajectories, and temporally aligned captions, while their answers span categorical labels, numerical values, multiple-choice options, time intervals, and free-form descriptions.

\FloatBarrier
\section{Reproducibility and Responsible Use}
\label{app:reproducibility_ethics}

\textbf{Reproducibility.}
To ensure reproducibility, Section~\ref{sec:method} details the HoliWorld pipeline, covering geometric reconstruction, static object perception, dynamic tracking, and supervision construction, alongside dataset composition and QA construction. Appendix~\ref{app:experimental_details} further specifies evaluation datasets, metric definitions, baseline adaptations, matching protocols, and ablation configurations.

\textbf{Responsible use.}
HoliWorld is designed for research on scene geometry and spatial-temporal understanding. Responsible data collection and release therefore require appropriate consent, review and redaction of identifying content, and compliance with dataset and model licenses. The pipeline should not be used for non-consensual surveillance or high-stakes decisions without additional safeguards and validation.

\end{document}